\documentclass[journal]{IEEEtran}
\usepackage{ifpdf}
\usepackage{multirow}
\usepackage{comment}

\usepackage{cite}
\ifCLASSINFOpdf
  \usepackage[pdftex]{graphicx}
\else
\fi
\usepackage{amsmath}
\usepackage{cleveref}
\usepackage{amssymb}
\usepackage{algorithmic}
\usepackage{array}
\usepackage{booktabs}
\newcommand{\ra}[1]{\renewcommand{\arraystretch}{#1}}
\usepackage{breqn}

\ifCLASSOPTIONcompsoc
  \usepackage[caption=false,font=normalsize,labelfont=sf,textfont=sf]{subfig}
\else
  \usepackage[caption=false,font=footnotesize]{subfig}
\fi
\begin{document}
\title{Localization Requirements for\\ Autonomous Vehicles}
\author{Tyler~G.~R.~Reid, 
	Sarah~E.~Houts, 
	Robert~Cammarata, 
	\\
	Graham~Mills, 
	Siddharth~Agarwal, 
	Ankit~Vora,
	and~Gaurav~Pandey% <-this % stops a space
	\thanks{T. G. R. Reid and G. Pandey are with
		Research and Advanced Engineering, Ford Motor Company, Palo Alto, CA 94304, e-mail: \{treid21, gpandey2\}@ford.com}% <-this % stops a space
	\thanks{S. E. Houts, and G. Mills are with
		Ford Autonomous Vehicles LLC, Palo Alto, CA 94304, e-mail: \{shouts, gmills47\}@ford.com}% <-this % stops a space
	\thanks{Robert Cammarata, S. Agarwal, and A. Vora are with
		 Ford Autonomous Vehicles LLC, Dearborn, MI 48124, e-mail: \{rcammar1, sagarw20, avora3\}@ford.com}% <-this % stops a space
	}%
\maketitle

\begin{abstract}
Autonomous vehicles require precise knowledge of their position and orientation in all weather and traffic conditions for path planning, perception, control, and general safe operation. Here we derive these requirements for autonomous vehicles based on first principles. We begin with the safety integrity level, defining the allowable probability of failure per hour of operation based on desired improvements on road safety today. This draws comparisons with the localization integrity levels required in aviation and rail where similar numbers are derived at 10\textsuperscript{-8}~probability of failure per hour of operation. We then define the geometry of the problem, where the aim is to maintain knowledge that the vehicle is within its lane and to determine what road level it is on. Longitudinal, lateral, and vertical localization error bounds (alert limits) and 95\% accuracy requirements are derived based on US road geometry standards (lane width, curvature, and vertical clearance) and allowable vehicle dimensions. For passenger vehicles operating on freeway roads, the result is a required lateral error bound of 0.57~m (0.20~m,~95\%), a longitudinal bound of 1.40~m (0.48~m,~95\%), a vertical bound of 1.30~m (0.43~m,~95\%), and an attitude bound in each direction of 1.50~deg (0.51~deg,~95\%). On local streets, the road geometry makes requirements more stringent where lateral and longitudinal error bounds of 0.29~m (0.10~m,~95\%) are needed with an orientation requirement of 0.50~deg (0.17~deg,~95\%).
\end{abstract}

\begin{IEEEkeywords}
Autonomous vehicles, automated driving, localization, positioning, requirements, safety, integrity
\end{IEEEkeywords}

% \fi
\IEEEpeerreviewmaketitle

%%%%%%%%%%%%%%%%%%%%%%%%%%%%%%%%%%%%%%%%%%%%%
%% Sections
%%%%%%%%%%%%%%%%%%%%%%%%%%%%%%%%%%%%%%%%%%%%%

\section{Introduction}
\label{Sec:Introduction} 

\begin{figure}
	\centering
	{\includegraphics[width=3.4in]{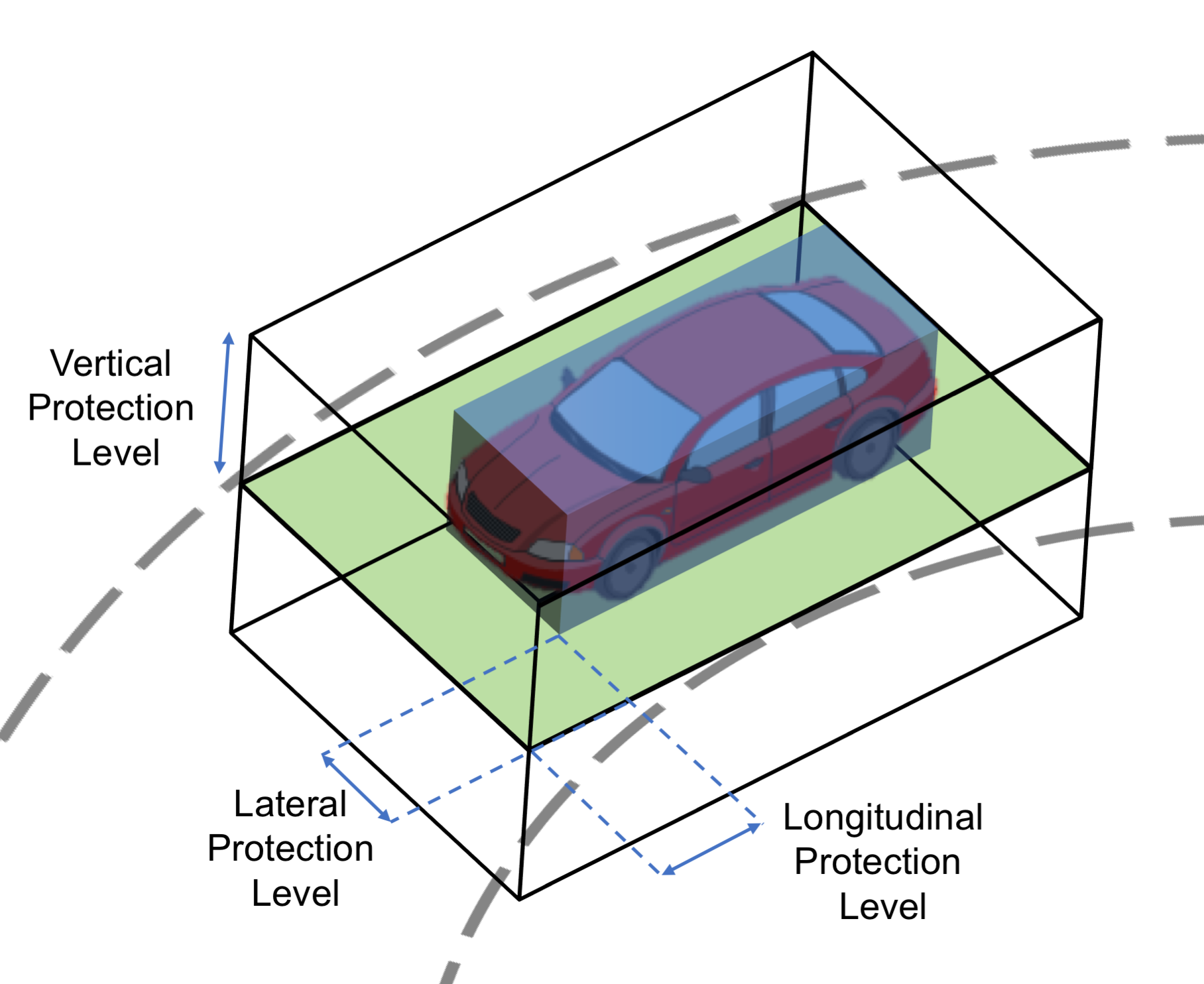}
		\caption{Definition of localization protection levels for automotive applications.}
		\label{fig:PL-def}}
\end{figure}

\IEEEPARstart{T}{he} challenge facing localization for autonomous systems in terms of required accuracy and reliability at scale is unprecedented. As will be shown, autonomous vehicles require decimeter-level positioning for highway operation and near-centimeter level for operation on local and residential streets. These requirements stem from one goal: ensure that the vehicle knows it is within its lane. Horizontally, this is broken down by lateral (side-to-side) and longitudinal (forward-backward) components. Vertically, the vehicle must know what road level it is on when located among multi-level roads. At any given time, the vehicle will have an estimate of its maximum position error in each direction. These are known as protection levels and are depicted in Figure \ref{fig:PL-def}. The maximum allowable protection levels in each direction to ensure safe operation are known as the alert limits. These alert limits are design variables; in our case they need to be sufficiently small to ensure that the vehicle stays within its lane at all times. If the protection level is larger than the alert limit at any given time, there is less certainty the vehicle will remain within its lane. These bounds will be shown to be a function of vehicle dimensions (length and width) along with road geometry standards (lane width and curvature). 

%Figure 1: Definition of localization protection levels for automotive applications.	

%Table 1: Evolution of localization accuracy in the last century [1]. 
\begin{table*}\centering
	\caption{Evolution of localization accuracy in the last century. Based on data from~\cite{Reid2017, weems1951, Dippy1946, Kelly1986, Lo2013, Stansell1968, Stansell1971, PARKINSON1995,  WilliamJ.HughesTechnicalCenterFederalAviationAdministrationGPS2011}. }
	\label{tab:accuracy-trend}
	\begin{tabular}{@{}lccccl@{}}\toprule
		System & Years Active & Horizontal Accuracy [m] & Latency &	Fix Type & Coverage\\ \midrule
		Celestial /  Chronometry & 1770 -- 1920 &  3,200 & Hours &	2D	& Global but not available when overcast\\ 
		LORAN-C & 1957 -- 2010 &  460 & None & 2D & North America, Europe, Pacific Rim\\
		Transit & 1964 -- 1996 &  25 & 30 -- 100 min & 2D & Global\\
		GPS & 1995 -- Present &  3 & None & 3D & Global\\
		\bottomrule
	\end{tabular}
\end{table*}

%Figure 2: Society of Automotive Engineers (SAE) levels of road vehicle autonomy (source: National Highway Traffic Safety Administration) 
\begin{figure*}
	\centering
	{\includegraphics[width=5.0in]{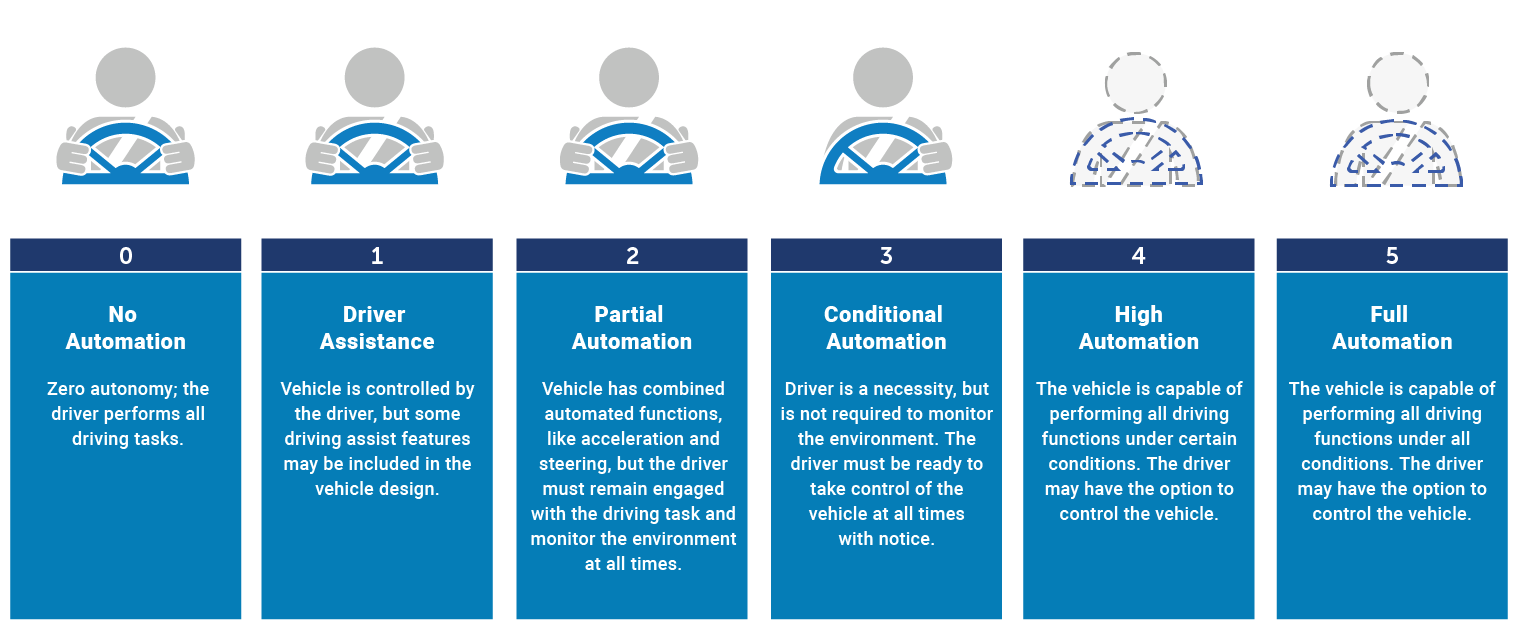}
		\caption{Society of Automotive Engineers (SAE) levels of road vehicle autonomy (source: National Highway Traffic Safety Administration).}
		\label{fig:sae-levels}}
\end{figure*}

%Figure 1a: Decade of the decimeter.  
\begin{figure}
	\centering
	{\includegraphics[width=3.5in]{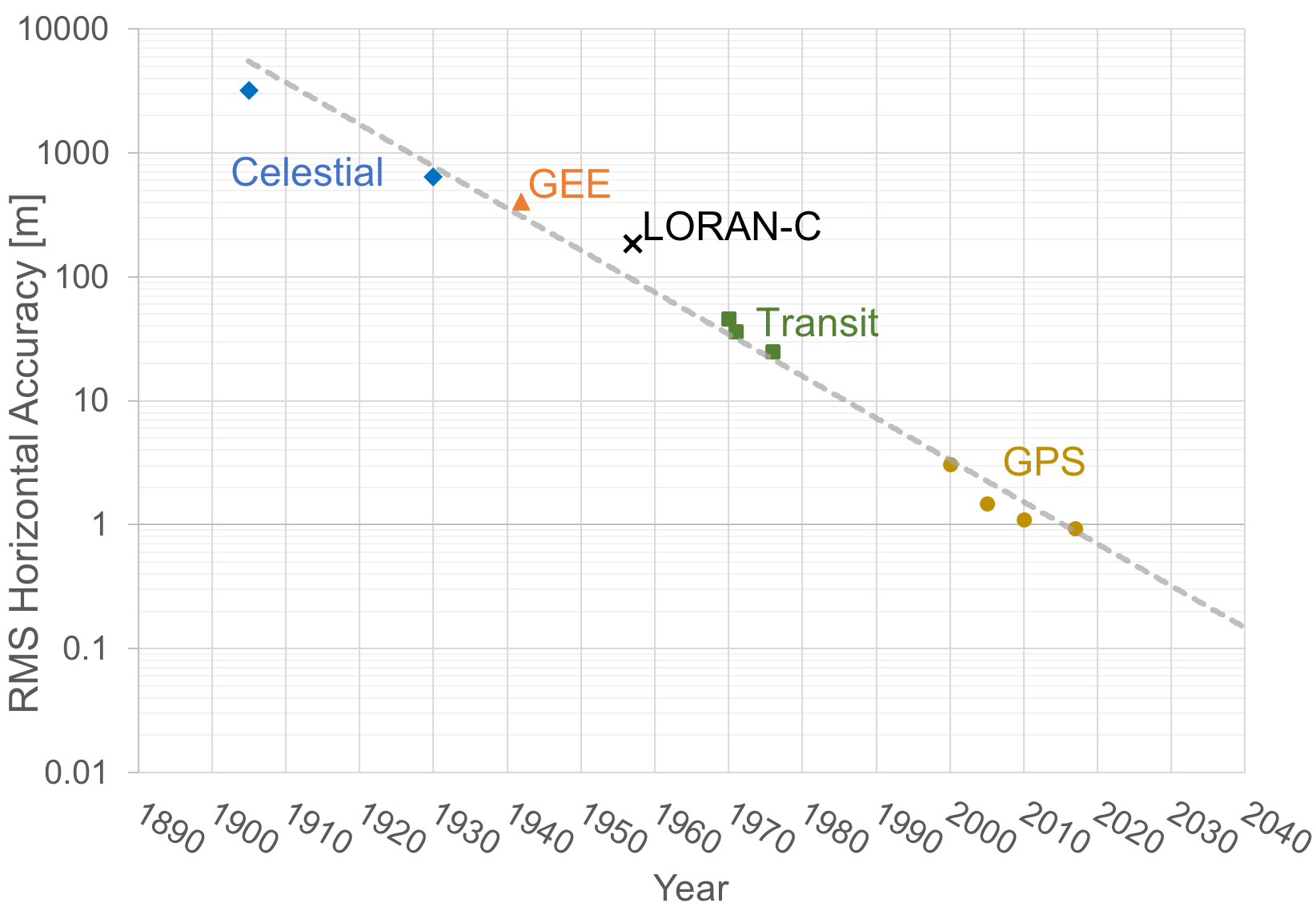}
		\caption{Historical trend in widely available RMS position accuracy from a variety of technologies. Compiled by the authors based on data from \cite{weems1951, Dippy1946, Kelly1986, Lo2013, Stansell1968, Stansell1971, PARKINSON1995,  WilliamJ.HughesTechnicalCenterFederalAviationAdministrationGPS2001, WilliamJ.HughesTechnicalCenterFederalAviationAdministrationGPS2006, WilliamJ.HughesTechnicalCenterFederalAviationAdministrationGPS2011, WilliamJ.HughesTechnicalCenterFederalAviationAdministrationGPS2017}.}
		\label{fig:decade-of-decimeter}}
\end{figure}

The challenge of decimeter location accuracy is put in perspective by Table \ref{tab:accuracy-trend} which shows the progress in localization throughout the last century~\cite{Reid2017}. In the early 1900s, the state of the art were the tools used for navigation at sea. This consisted of a sextant to measure latitude by the stars and a precise mechanical clock known as a marine chronometer to measure longitude. Combined, these sensors gave rise to approximately 3~km of position accuracy at sea~\cite{weems1951}. The Second World War accelerated the use of radio in navigation to support emerging aviation operations. Air navigation was needed in all weather in real-time. Land-based radio beacons were developed which gave rise to approximately 500~meters of accuracy but this required proximity to this infrastructure's limited range~\cite{Dippy1946,Kelly1986, Lo2013}. In the 1960s, the first satellite navigation system, Transit, came online. Operated by the US Navy, Transit offered 25~meters of accuracy, supporting the localization requirements of Polaris ballistic missile submarines~\cite{Stansell1968, Stansell1971, PARKINSON1995}. This system had only a handful of satellites in low Earth orbit, resulting in sometimes an hour or more to obtain a position fix. In the 1990s, the Global Positioning System (GPS) came online, now offering 1--5~meters of accuracy in open skies everywhere on Earth in real-time~\cite{WilliamJ.HughesTechnicalCenterFederalAviationAdministrationGPS2017, VanDiggelen2015}. Figure \ref{fig:decade-of-decimeter} shows this progression as a clear trend in the last century: an order of magnitude increase in localization accuracy every 30~years. This predicts the 2020s to be the first decade of the decimeter. As will be shown here, we are well poised, as this trend is towards the needs of autonomous vehicle operation. 

Figure \ref{fig:sae-levels} shows the Society of Automotive Engineers (SAE) levels of vehicle autonomy~\cite{SAEInternational2018}. Level 0 represents no automation where the driver is responsible for all aspects of driving and exemplifies most vehicles up to approximately the 1970s. Level 1 represents some driver assistance features for either braking / throttle or steering but the driver is still expected to monitor and control the vehicle. This includes features like cruise control and Anti-lock Braking Systems (ABS). Level 2 is partial automation, where the human driver is responsible for monitoring the scene and the system is responsible for some dynamic driving tasks including lateral and longitudinal motion (steering, propulsion, and braking). The human driver must be ready to take over dynamic driving tasks immediately when the driver determines the system is incapable. Examples of level 2 systems are Tesla's Autopilot released in 2014 and General Motor's Super Cruise released in 2017~\cite{Hughes2018, Ackerman2017}. Level 3 is conditional automation, meaning that in certain circumstances, the vehicle is within its Operational Design Domain (ODD). The system is responsible for monitoring the scene and dynamic driving tasks including lateral and longitudinal motion (steering, propulsion, and braking). The human driver must be ready to take over dynamic driving tasks within a defined time when the system determines it's incapable or the vehicle is outside of its ODD and notifies the driver. Level 4 is highly automated where the system can perform all dynamic driving tasks within its ODD. This mode of operation maybe geofenced to areas with appropriate supporting infrastructure (e.g. maps and connectivity) and may further be restricted to certain weather conditions. In 2018, this represents largely research vehicles such as Waymo's self-driving minivan which is gearing towards initial ride-sharing service~\cite{Harris2018}. Level 5 is full automation, where the vehicle is capable of performing all dynamic driving tasks in all areas and under all conditions.

Positioning accuracy requirements for connected vehicle (V2X) applications have been previously broken down by Basnayake et al.~\cite{Basnayake2010} into the categories of which-road ($<$5~meters), which-lane ($<$1.5~meters), and where-in-lane ($<$1.0~meter) based on the desired function or operation. The National Highway Safety Administration (NHTSA) has, as part of its Federal Motor Vehicle Safety Standards in V2V Communications, determined that position must be reported to an accuracy of 1.5 meters (1$\sigma$ or 68\%) as this is tentatively believed to provide lane-level information for safety applications~\cite{NationalHighwayTrafficSafetyAdministration2017}. Vehicle positioning requirements were further developed by Stephenson~\cite{Stephenson2016} who explored the Required Navigation Performance (RNP) for Advanced Driver Assistance Systems (ADAS) and automated driving. Stephenson added the functional category of active vehicle control requiring an accuracy better than 0.1~meters. 

\begin{figure}
	\centering
	{\includegraphics[width=3.5in]{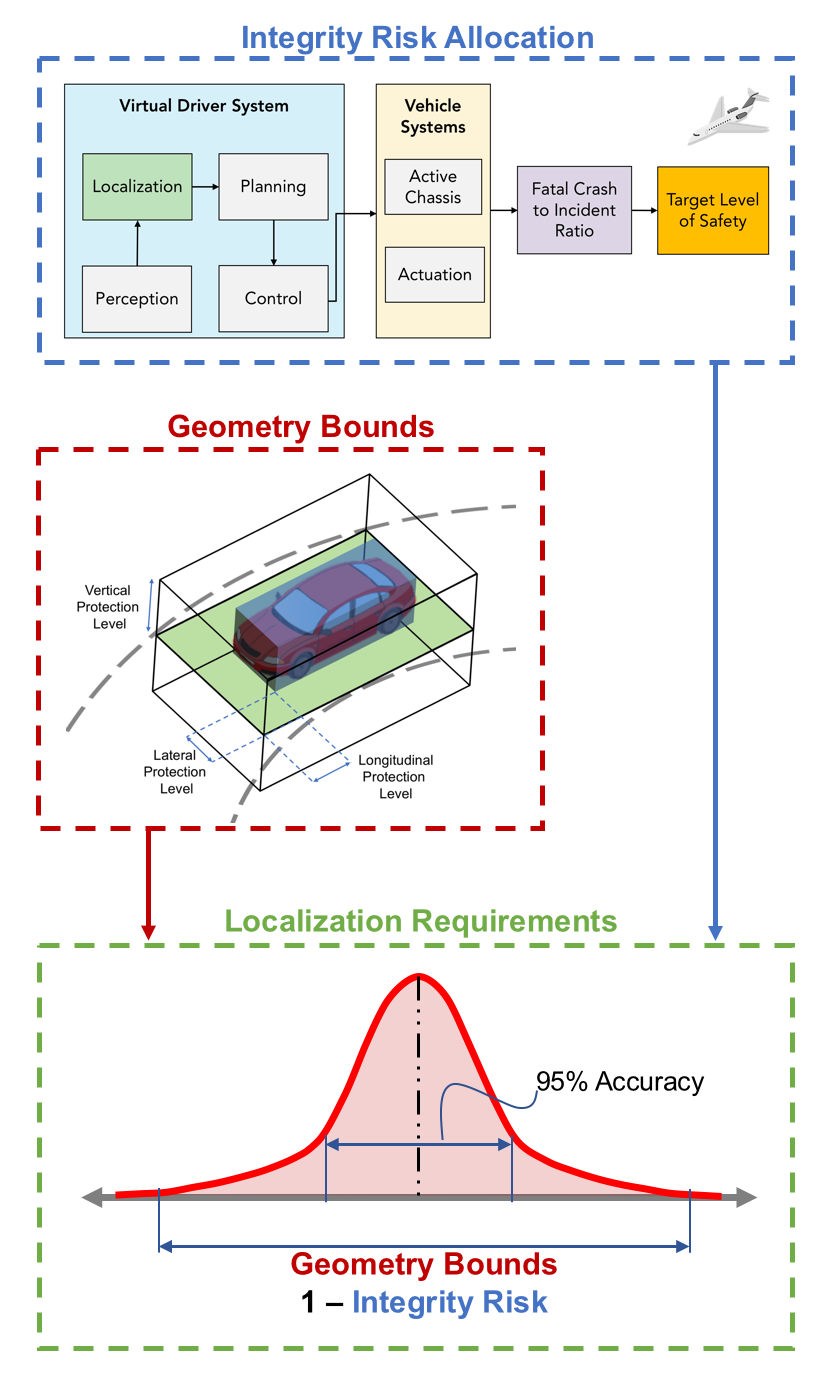}
		\caption{Our approach to developing localization requirements for autonomous vehicles. We derive system integrity risk allocation based on a target level of safety. This risk budget is then distributed throughout the autonomous vehicle system, following methodologies developed in civil aviation. We define the geometry of the problem to establish positioning bounds based on vehicle dimensions and road geometry. Combining these defines the desired distribution of our position errors and the localization requirements. }
		\label{fig:outline}}
\end{figure}

In this paper, we focus on full autonomous operation which requires the accuracy needed for active control. In this context, `autonomous vehicle' will refer to level 3+ systems, though some of what is shown here can also be applied to level 2. The process followed to develop these requirements is outlined in Figure \ref{fig:outline}. We begin with defining the statistics of the problem by establishing a target level of safety. Following methodologies developed in civil aviation, the target level of safety is used to allocate appropriate integrity risk to each element of the system including localization. Next, we define the geometry of the problem to establish positioning bounds. This will be shown to be a function of road geometry standards such as lane widths and road curvature along with permissible vehicle dimensions. This analysis will focus on road and passenger vehicle standards in the United States, though the same principles can be applied to other regions and vehicle types. 

%Figure 3a: Outline

%Figure 3: Relationship between protection level, alert limit, availability, and different operations. (a) Shows the case where PL < AL along with examples of nominal, misleading, and hazardous misleading information. (b) Shows the case where AL < PL resulting in no availability. 
\begin{figure}
	\centering
	\subfloat[Case I: $PL < AL$ along with examples of nominal, misleading, and hazardous misleading information.]{\includegraphics[width=3.0in]{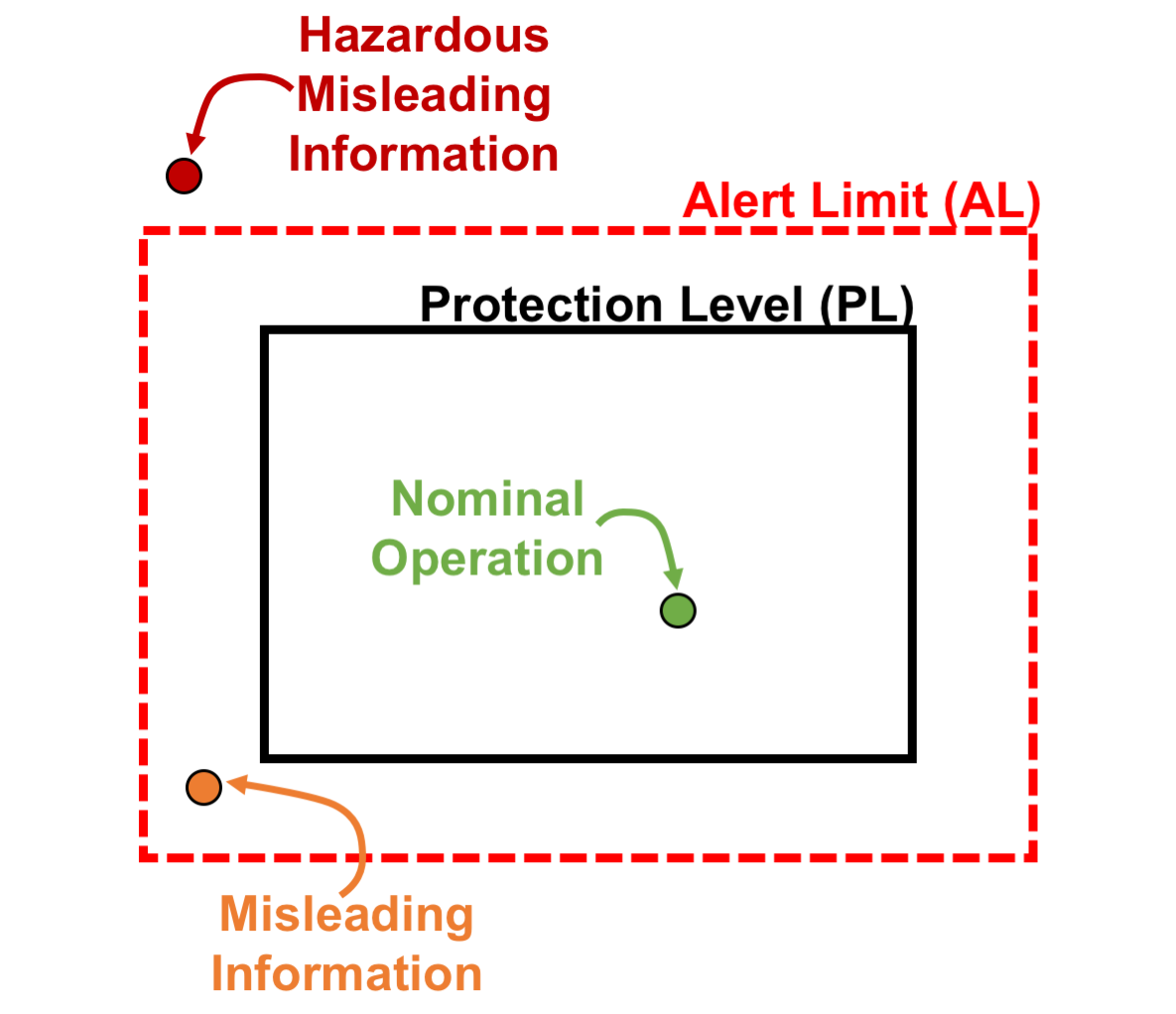}%
		\label{fig_first_case}}
	\hfil
	\subfloat[Case II: $AL < PL$ resulting in no availability.]{\includegraphics[width=3.0in]{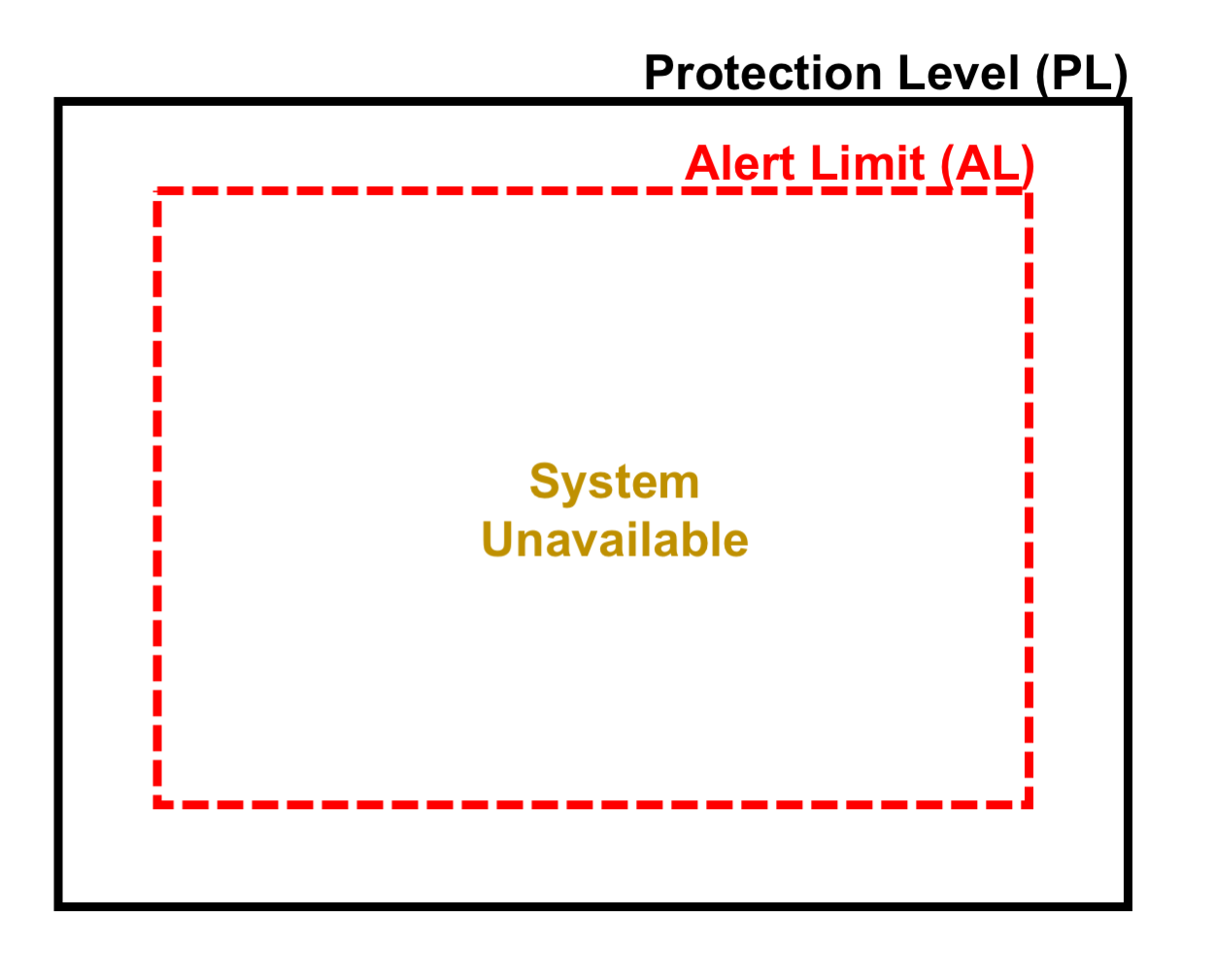}%
		\label{fig_second_case}}
	\caption{Relationship between protection level, alert limit, availability, and different operations. 
	}
	\label{fig:hmi-avail}
\end{figure}

In safety critical localization systems, the instantaneous estimate of the maximum possible error in position is known as the protection level. Figure \ref{fig:PL-def} shows our definition of the lateral, longitudinal, and vertical protection levels around the vehicle. We define this as a box as this is a logical breakdown for road vehicles, though other forms based on ellipsoids have been proposed~\cite{Feng2018}. Some sensors and systems are better at providing lateral information such as cameras which can recognize lane lines and other types of sensors that can provide longitudinal information such as wheel odometry. The hard bounds on allowable lateral, longitudinal, and vertical protection levels are known as alert limits and are design choices which are dictated by geometry. We will choose these such that we always know we are within the lane. If our protection level is larger than our alert limit we cannot guarantee we are within the lane. Together, the desired integrity level and geometric bounds define the requirements of localization as a system. 

Safety-critical localization systems specify their performance in terms of accuracy, integrity, and availability. Availability is a measure of how often our protection levels are larger than our alert limits. If we are always within the maximum permissible error (alert limits), we have 100\% availability. If we do so only some of the time, we have only limited availability. As a system, when operating in autonomous mode, localization must always be available. On the flip side, localization availability could be one of the metrics used to determine if the vehicle is within its Operational Design Domain to enable autonomous features. This could be a function of where high definition maps or other forms of supporting infrastructure are present. 

\begin{figure}
	\centering
	{\includegraphics[width=3.3in]{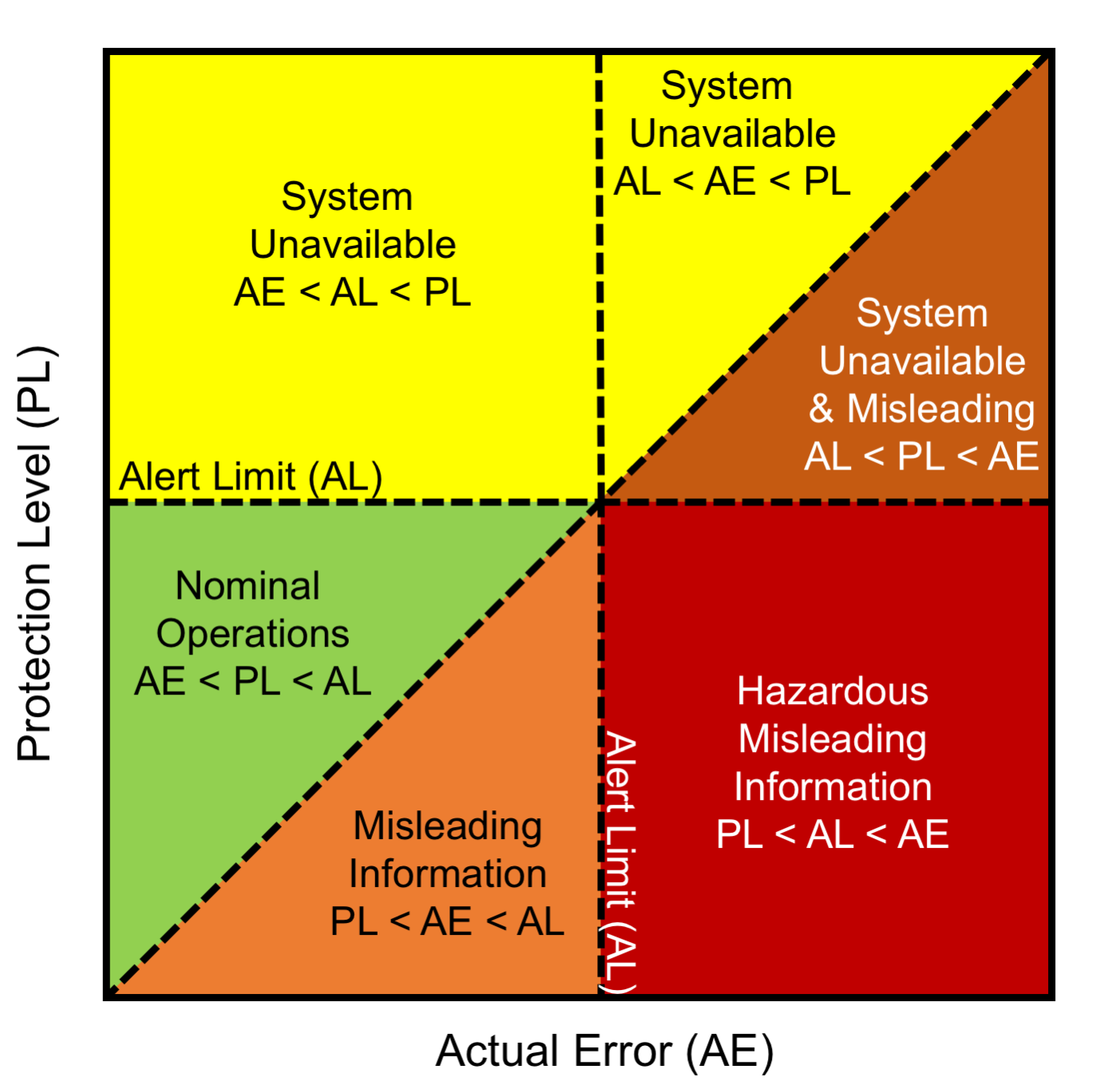}
		\caption{The Stanford Diagram. This shows the relationship between actual error, protection level, and alert limit.}
		\label{fig:stanford-diag}}
\end{figure}

Integrity describes how often our true error is outside of our estimate of maximum possible error or protection level. Outside of this and hazardous information is being fed to the vehicle's decision making and control systems. This is the probability of system failure, usually specified as probability of failure per hour of operation. We will derive the requirement on integrity based on the desired level of road safety. This will specify a level which represents improvement in road safety today by drawing comparisons to safety levels achieved in commercial aviation. We will also tie this integrity number to existing design safety standards in automotive and other industries. Figures \ref{fig:hmi-avail} and \ref{fig:stanford-diag} summarize the relationship between protection level, alert limit, availability, misleading information, and hazardous misleading information. These will be developed in more detail throughout the text. 

Accuracy is described by the typical performance of the system, usually measured at the 95th percentile. This is a manifestation of the desired statistical distribution of localization errors, where alert limits usually define the maximum allowable error for operation to the desired integrity level which define the tails of the statistical distribution. Hence, accuracy is a measure of nominal performance and integrity a measure of the limits. This relationship is shown in Figure \ref{fig:outline}.

%Figure 4: The Stanford Diagram. This shows the relationship between actual error, protection level, and alert limit. 

%Figure 5: Example of Ford’s research autonomous vehicle platform sensors used in localization and perception.
\begin{figure}
	\centering
	{\includegraphics[width=3.5in]{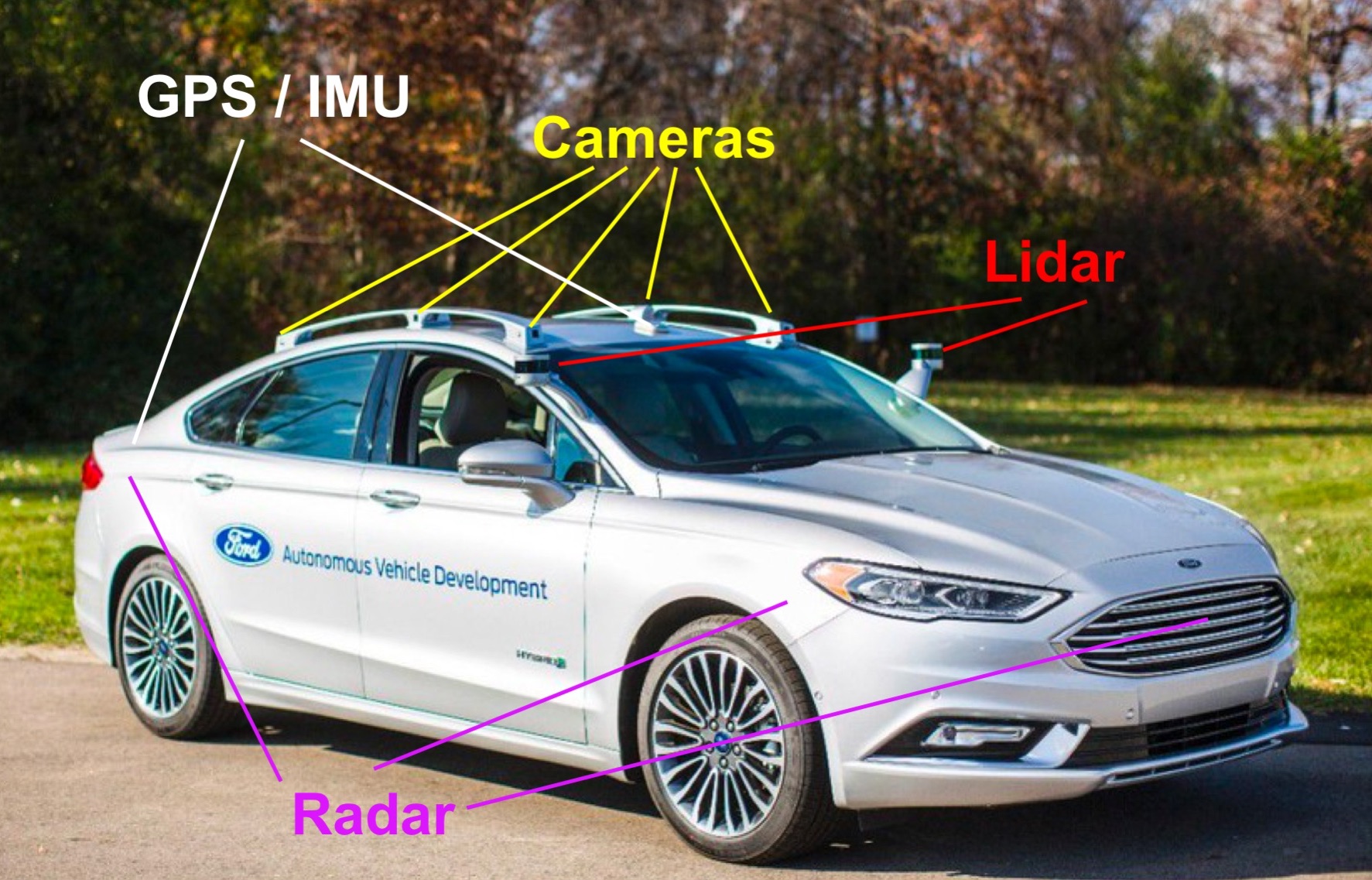}
		\caption{Example of Ford's research autonomous vehicle platform sensors used in localization and perception.}
		\label{fig:ford-av}}
\end{figure}

At present, there is no one localization technology that can meet the requirements presented here for safe operation in all weather, road, and traffic scenarios. Today, Ford Motor Company's autonomous vehicle research platform makes use of a complex sensor fusion strategy for localization and perception as shown in Figure \ref{fig:ford-av}. This includes LiDAR, radar, cameras, a Global Navigation Satellite System (GNSS) receiver, and an Inertial Measurement Unit (IMU). Each of these are a system in itself, and hence localization is a system of systems. Since no one technology can achieve the requirements in all scenarios, this will require multi-modal sensing and their intelligent combination to achieve the integrity levels needed for safe operation. 

Though the sensor strategy shown in Figure \ref{fig:ford-av} can meet many of the requirements that will be developed here in a research setting, the sensor costs are such that many challenges lie ahead on the path to production. Decimeter accuracy will require a trade-off between onboard sensors, compute resources / storage, and supporting infrastructure such as high definition maps and GNSS corrections. Such infrastructure may also limit where self-driving features are enabled due to availability of precise localization. Ultimately, precise vehicle location is only useful in context and hence a-priori maps are a major piece of the puzzle. We will limit the scope of this paper to localization requirements as mapping is a vast and complex area in its own right. We will touch on mapping requirements as they relate to localization.

\section{Integrity}
\label{Sec:Integrity} 

In 2016, there were 34,439 fatal car crashes in the United States, resulting in 37,461 fatalities~\cite{InsuranceInstituteforHighwaySafety, U.S.DepartmentofTransportation2017}. On a per mile basis, this is 1.18 fatalities per 100 million miles traveled (1.18$\times$10\textsuperscript{-8}~fatalities~/~mile)~\cite{U.S.DepartmentofTransportation2017}. The cause of vehicle crashes are estimated as follows: 94\%($\pm$2.2\%) from human driver errors; 2\%($\pm$0.7\%) from electrical and mechanical component failures; 2\%($\pm$1.3\%) from environmental factors contributing to slick (low $\mu$) roads such as water, ice, snow, etc.; and 2\%($\pm$1.4\%) from unknown reasons~\cite{Singh2015}. In total, road fatalities account for 95\% of all transportation related fatalities in the United States where 2\% come from rail, 2\% from water transport, and 1\% from aviation~\cite{U.S.DepartmentofTransportation2017}. Though the number of fatalities per road mile has decreased by five-fold since the 1960s, it has remained relatively constant over the last decade~\cite{U.S.DepartmentofTransportation2017, Oster2013}. 

\begin{figure}[h]
	\centering
	{\includegraphics[width=3.6in]{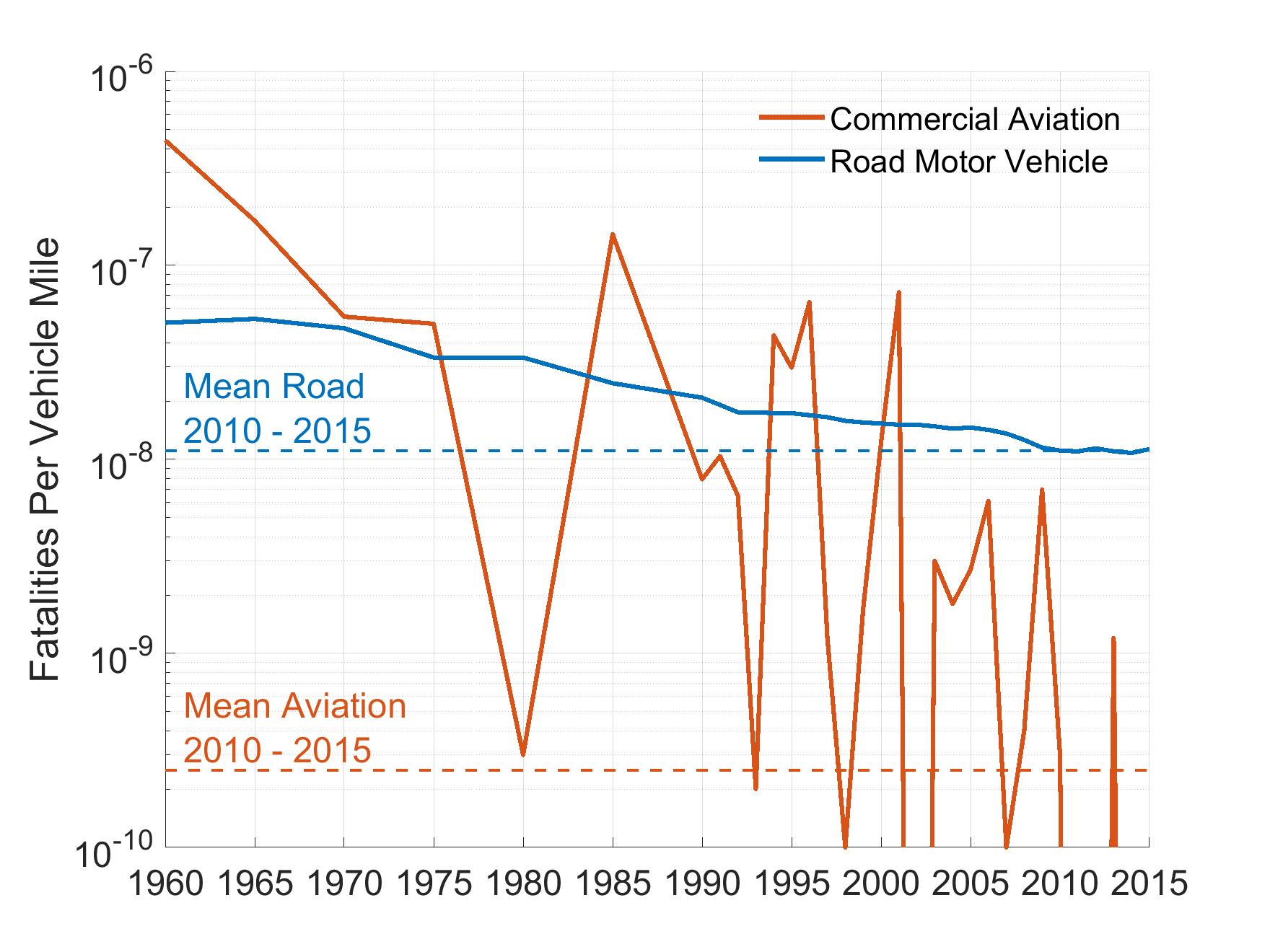}
		\caption{Fatalities per vehicle mile 1960 -- 2015 for road vehicles and commercial aviation in the US. Based on data from~\cite{U.S.DepartmentofTransportation2017}.}
		\label{fig:fatalies-per-mile}}
\end{figure}

Figure \ref{fig:fatalies-per-mile} compares fatalities on the road with commercial aviation on a per vehicle mile basis between 1960 -- 2015 in the US. This shows that between 2010 -- 2015, commercial aviation averaged 2.50$\times$10\textsuperscript{-10} fatalities per mile, making air travel nearly two orders of magnitude safer than road travel when using this metric. Improvements in road safety are being proposed through automation, where sensors, silicon, and software are being combined into a virtual driver system which aims to perform many of the tasks of the human driver today. The first generation of commercially available virtual driver systems must be safer than the human drivers they aim to replace if they are to be socially accepted.  Vehicle component failures are currently responsible for only 2\% of crashes~\cite{Singh2015}. If this same metric is upheld for virtual driver systems which aim to replace the human factors representing 94\% of crash causation today, then we strive to achieve roughly two orders of magnitude improvement in road safety. This brings the necessary system requirements for autonomous road vehicles up to the levels mandated in civil aviation, an industry known for its safety achieved through strict training for air and ground crew, supporting infrastructure, and well-developed standards. 

% Figure, fatalities per mile. 

% Figure, accident to incident ratio. 
\begin{figure}[h]
	\centering
	{\includegraphics[width=3.6in]{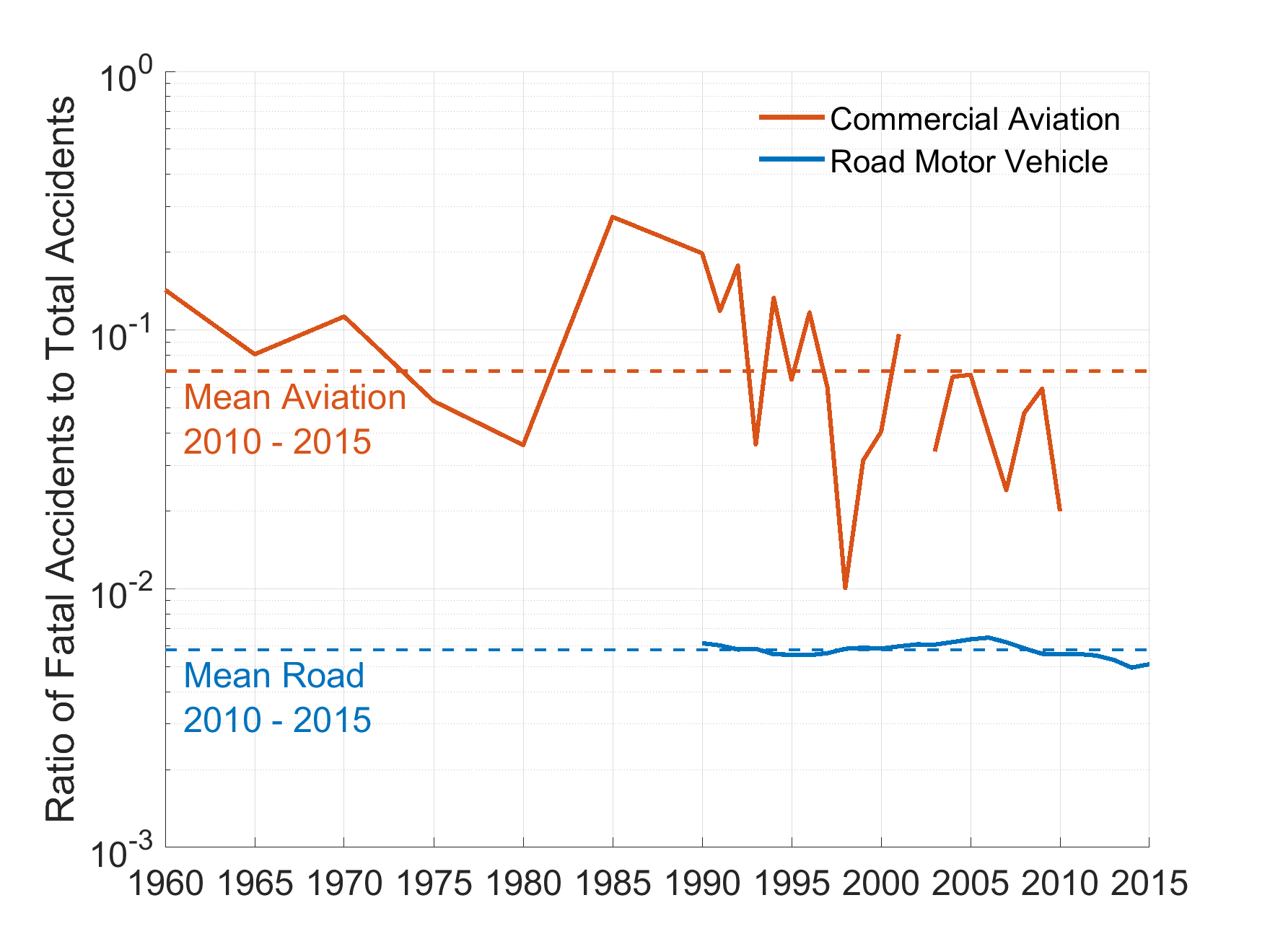}
		\caption{Ratio of fatal accidents (crashes) to total reported accidents (crashes) between 1960 -- 2015 for road vehicles and commercial aviation in the US. Based on data from~\cite{U.S.DepartmentofTransportation2017} where data for road vehicles was only available from 1990 onward. In some recent years, commercial aviation saw no fatal accidents and hence the appearance of missing data.}
		\label{fig:accident-incident-ratio}}
\end{figure}

% Figure 6: Fault tree allocation for localization as part of the autonomous vehicle system.  
\begin{figure*}[h]
	\centering
	{\includegraphics[width=7.0in]{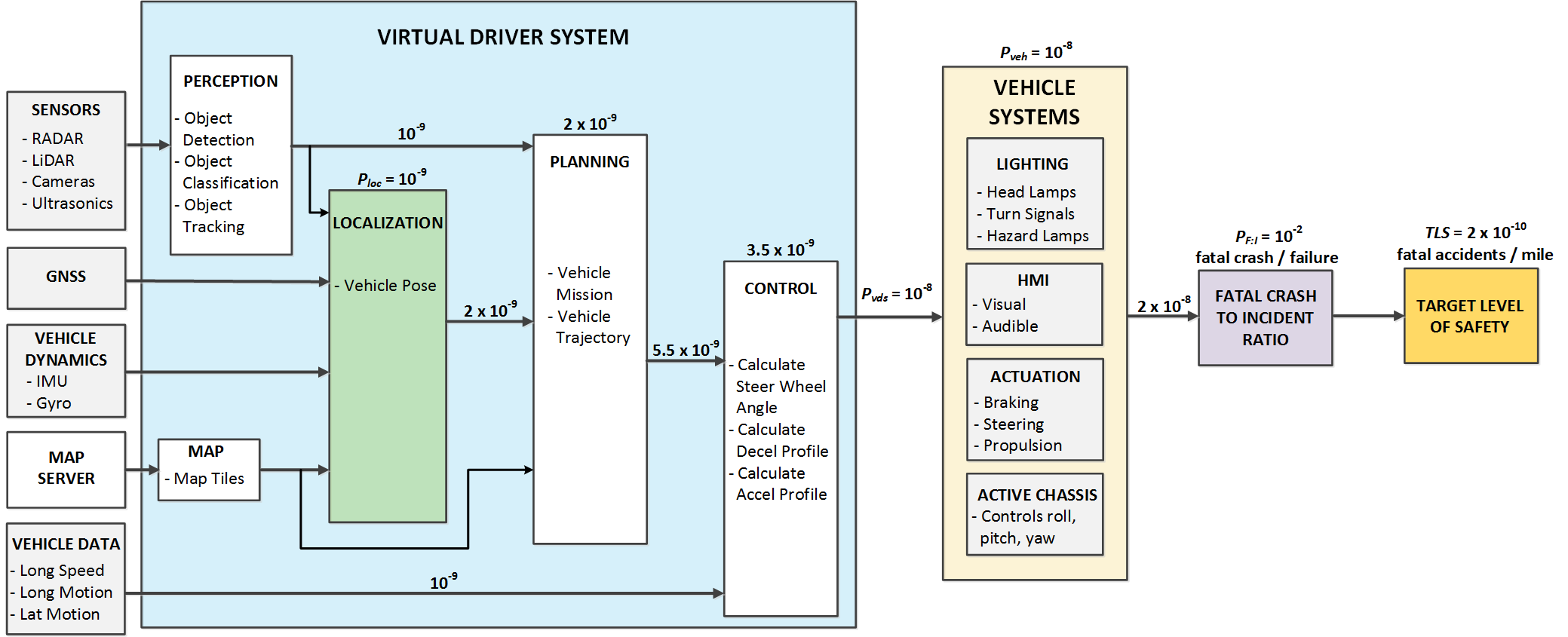}
		\caption{Virtual driver system integrity risk allocation. Unless otherwise indicated, values are given as failures per mile. This diagram makes certain assumptions about how faults cascade through the system. For example, failures in localization output are assumed to lead directly to failures in planning though planning can also have its own independent failures as well. The target level of safety is derived based on numbers achieved in practice in aviation between 2010 - 2015. The fatal crash to incident ratio is based on historical road data from~\cite{U.S.DepartmentofTransportation2017} (see Figure \ref{fig:accident-incident-ratio}). } 
		\label{fig:fault-tree}}
\end{figure*}

Our starting point for this analysis will be the goal of a one hundred times improvement in road safety through the virtual driver system. This is the level of commercial aviation today at 2.50$\times$10\textsuperscript{-10} fatalities per mile and hence will be our target. To put this in units of fatal crashes per road vehicle mile, we must divide by the ratio of fatalities to fatal crashes which in 2016 was 37,461$/$34,439~=~1.09~\cite{U.S.DepartmentofTransportation2017}. Being conservative, this leads to a target level of safety of $TLS=$ 2$\times$10\textsuperscript{-10} fatal crashes per vehicle mile. To translate this into system level requirements, we must examine both historical data and the autonomous vehicle architecture, an approach mirroring that taken in civil aviation~\cite{Busch1985}. 

Figure \ref{fig:fault-tree} shows the system breakdown of the autonomous vehicle including the virtual driver and other vehicle modules. Furthermore, it shows one possible of allocation of integrity risk across individual elements with an end target level of safety of $TLS=$ 2$\times$10\textsuperscript{-10}~fatal crashes~/~vehicle mile. To arrive at this distribution, we must work backwards from the $TLS$. We must first account for the fact that not every malfunction will directly lead to a hazard that will cause a fatal crash. Some failures may only lead to a lane departure or minor crash. In aviation, this fatal accident to incident ratio $P_{F:I} $ is taken as 1:10~\cite{Kelly1994, Roturier2001}, and Figure \ref{fig:accident-incident-ratio} shows why. This plot shows the yearly ratio of fatal to total reported accidents for commercial aviation and road transport. Aviation is shown to be 1:14 and automotive 1:172. This is not unexpected, due to the higher speed of operation of aircraft and consequently, the severity of crashes. Based on this historical data, we have conservatively chosen the automotive fatal crash to incident ratio as $P_{F:I} =$ 10\textsuperscript{-2}~fatal~crashes~/~failure, meaning one fatal crash for every one hundred incidents where an incident could be seen as a lane departure or minor crash. 

Our chosen fatal crash to incident ratio is conservative, since this number represents the ratio of fatal crashes to all police reported crashes, not necessarily to all incidents. The crash data in commercial aviation is generally reliable due to the strict reporting requirements. In automotive, this is less so, as not all minor crashes are reported. It is estimated that 60\% of property-damage-only crashes and 24\% of all injury crashes are not reported to the police~\cite{Blincoe2015}. Furthermore, the data supports that the 1 fatal crash for every 100 police reported crashes is consistent between functional road types, e.g. interstate and local urban roads, at least to within a factor of 2 -- 3~\cite{OregonDepartmentofTransportation2013, OregonDepartmentofTransportation2014}. Hence, we believe we are within an order of magnitude with this estimate for the majority of driving scenarios. 

The next step is to allocate the acceptable level of integrity risk to vehicle systems $P_{veh}$ and to the virtual driver system $P_{vds}$. These are related to the target level of safety $TLS$ and fatal crash to incident ratio $P_{F:I}$ as follows: 
\begin{dmath} \label{eq:p-p-p-TLS-no-numbers}
	TLS = P_{F:I} \left(  P_{veh} + P_{vds} \right)
\end{dmath}
To reach our $TLS=$ 2$\times$10\textsuperscript{-10}~fatal crashes~/~vehicle mile, we will allocate equal integrity risk to both the virtual driver and vehicle systems. This works out to $P_{vds} = P_{veh}$ = 10\textsuperscript{-8}~failures~/~mile and is reflected in Figure \ref{fig:fault-tree}. For clarity, plugging in these values into equation (\ref{eq:p-p-p-TLS-no-numbers}) gives the desired result: 
\begin{dmath} \label{eq:p-p-p-TLS}
	TLS = P_{F:I} \left(  P_{veh} + P_{vds} \right) = 10^{-2} \frac{\text{fatal crashes}}{\text{failure}} \left(  10^{-8} +  10^{-8} \right) \frac{\text{failure}}{\text{mile}}  = 2 \times 10^{-10}~\text{fatal crashes}/\text{mile}
\end{dmath}

Examination of historical data reveals that vehicle system failure rates $P_{veh}$ are very nearly 10\textsuperscript{-8}~failures~/~mile today. This can be estimated through police reported crashes, NHTSA estimates of crash causation due to vehicle systems, and the number of vehicle miles driven. On average, between 2010 -- 2015 there was 5,800,000 police reported crashes per year on US roads~\cite{U.S.DepartmentofTransportation2017}. These crashes are the rounded sum of fatal crashes, an actual count from the Fatality Analysis Reporting System, injury crashes, and property damage only crashes, which are estimates from the National Automotive Sampling System-General Estimates System~\cite{U.S.DepartmentofTransportation2017}. Currently, NHTSA estimates 2\%($\pm$0.7\%) of these crashes to be caused by vehicle systems~\cite{Singh2015}. Using this, along with the fact that 3,005,829,000,000 miles were driven on average in the US between 2010 -- 2015, we can obtain an estimate of historical failure rate $\hat{P}_{veh}$: 
\begin{dmath} \label{eq:veh-integrity}
\hat{P}_{veh} = \frac{5,800,000~\text{crashes}}{3,005,829,000,000~\text{miles}} \times 2\%  = 3.8 \times10^{-8}~\text{failures}/\text{mile}
\end{dmath}
Though there are sources of uncertainty in each of the values used above, this shows that vehicle systems are at the proposed order of magnitude, indicating that 10\textsuperscript{-8}~failures~/~mile can likely be attained if it is not already. 

Achieving the desired integrity risk for the virtual driver system $P_{vds}$ will require a closer look at its subsystems. In our case, we are focused on localization. Localization will need to have a lower probability of failure since it feeds other elements of the virtual driver system. This includes hardware and software failures within perception, localization, planning, and control. Figure \ref{fig:fault-tree} shows the internal elements of the virtual driver system and the importance of localization within the system. The output of localization is an input to planning, and the output of planning is the input to control, therefore failures in localization propagate downstream. One possible allocation of integrity risk to all virtual driver subsystems is shown in Figure \ref{fig:fault-tree} where localization is targeted at $P_{loc} =$ 10\textsuperscript{-9}~failures~/~mile. This diagram assumes that failures at any given point downstream are a combination of upstream input failures along with failures of the given subsystem. $P_{loc}$ is typically referred to as the probability of the localization system outputting hazardous misleading information (as described in Section \ref{Sec:Introduction}). To get this in terms of failures per hour of operation (a more common unit), we first need to determine the vehicle speed range. Maximum speed limits in the US are found in Texas at 85~mph (137~km/h). On the lower side, we will consider the minimum speed at which airbags will deploy, which corresponds to 10~mph (16~km/h)~\cite{Wood2014}. These speeds give the following range: 
\begin{dmath} \label{eq:prob-fail-hour}
P_{loc} = 10^{-9}\frac{\text{failures}}{\text{mile}} \times \left( 10 - 85\right)  \frac{\text{mile}}{\text{hour}} \approx 10^{-8} \frac{\text{failures}}{\text{hour}}
\end{dmath}

In this analysis of the localization system, we will examine an allowable integrity risk of 10\textsuperscript{-8}~failures~/~hour of operation. This is the requirement on the localization system as a whole, which itself may be comprised of several subsystems, sensors, and independent localization algorithms based on GNSS, IMU, cameras, LiDAR, maps, and other elements. This is the number that must be achieved in all weather and traffic scenarios where the vehicle intends operation. The gold standard from ISO 26262 for automotive functional safety is 10 Failures In Time (FIT) which corresponds to 10 failures in one billion hours of operation or 10\textsuperscript{-8}~failures~/~hour of operation. This aligns with our intended target. This is Automotive Safety Integrity Level (ASIL) D, the highest standard for current automobiles. Though the error distribution of the localization system may not be Gaussian, when thinking of this in Gaussian terms, (1~--~10\textsuperscript{-8}) is 99.999999\% or approximately 5.73$\sigma$. 

% Table 2: Localization requirements for maritime, aviation, and rail.
\begin{table}\centering
	\caption{Localization requirements for maritime, aviation, and rail.} 
	\label{tab:loc-req-mar-avia-rail}
	\ra{1.6}
	\begin{tabular}{m{0.75in}m{0.65in}>{\centering\arraybackslash}m{0.3in}>{\centering\arraybackslash}m{0.25in}>{\centering\arraybackslash}m{0.55in}}\toprule
		Transport Mode	&	Operation	&	Accuracy (95\%) [m]	&	Alert Limit [m]	&	Probability of Failure \\ 
		\midrule
		\multirow{3}{*}{Maritime \cite{InternationalMaritimeOrganization2002}} & Open Ocean, Coastal & 10\textsuperscript{**} & 25\textsuperscript{**} & 10\textsuperscript{-5} / 3 h \\ 
		& Port, \mbox{Hydrography}, Drilling & 1\textsuperscript{**} & 2.5\textsuperscript{**} & 10\textsuperscript{-5} / 3 h \\
		\hline
		
		\multirow{3}{*}{Aviation \cite{Roturier2001, Speidel2013}} & LPV 200 Airport Approach & 4\textsuperscript{*} & 35\textsuperscript{*} & 10\textsuperscript{-7} / 150 s \\ 
		& CAT~II~/~III Instrument Landing & 2.9\textsuperscript{*} & 5\textsuperscript{*} & 10\textsuperscript{-9} / 150 s  \\
		\hline
		
		\multirow{3}{*}{Rail \cite{Marais2017}} & Train control & - & 20\textsuperscript{**} & 10\textsuperscript{-9} / h \\ 
		& Parallel track discrimination & - & 2.5\textsuperscript{**} & 10\textsuperscript{-9} / h \\
		
		\bottomrule
	\end{tabular}
	\raggedright
	\\	
	\rule{0pt}{2ex}  
	*Vertical \\
	**Horizontal
\end{table}

The localization requirements in maritime~\cite{InternationalMaritimeOrganization2002}, aviation~\cite{Roturier2001, Speidel2013, Kelly1994}, and rail~\cite{Marais2017, Filip2008} for specific operations are given in Table \ref{tab:loc-req-mar-avia-rail} for comparison. Here, we specify the 95\% localization accuracy, the alert limit which is the hard bound on position error to ensure safe operation, and the acceptable probability of system failure or integrity risk. In aviation, the operations given correspond to airport precision approach and landing. The timescale associated with probability of failure of 150~seconds corresponds to the typical time this operation takes. Localizer Performance with Vertical guidance (LPV~200) gets the aircraft down to a decision height of 200~feet (61~m) above the runway where the pilot can decide to either land the aircraft or fly around and make another approach. CAT~II~/~III is full instrument landing and hence the two orders of magnitude difference in acceptable failure rate since the system is fully automated. Maritime operations happen at lower speeds and is why integrity is specified over 3~hours. However, it can be shown that 10\textsuperscript{-5}~failures~per~3~hours is roughly equivalent to aircraft precision approach requirements at 10\textsuperscript{-7}~failures~per~150~seconds~\cite{Reid2016}. Rail has separate along-track and cross-track requirements. Along-track requirements describe where trains are along a given track known as train control. Cross-track requirements are needed to distinguish which track the train is on known as parallel track discrimination. Track discrimination requirements are most strict since the inter-track spacing is tighter than the spacing kept between trains on the same line. Both rail operations require a failure rate of 10\textsuperscript{-9}~failures~/~hour of operation. For the virtual driver localization system, we are aiming for 10\textsuperscript{-8}~failures~/~hour or better since the virtual driver system itself will be designed to ASIL-D standards. Though this target integrity level has precedence in other transportation industries, the required alert limits do not. We will show in the coming sections that alert limits needed for road vehicles is on the order of decimeters, an order of magnitude smaller than anything in Table \ref{tab:loc-req-mar-avia-rail}. 

%Table 3: Typical characterization of safety risk based on data from [20]–[24].
\begin{table}\centering
	\caption{Typical characterization of safety risk based on data from~\cite{Verhulst2013, Verhulst2013a, Baufreton2010, Blanquart2012, Machrouh2012}.} 
	\label{tab:sil-meaning}
	%\ra{1.3}
	\begin{tabular}{@{}lcl@{}}\toprule
		Category & Safety & Consequence of Failure\\
		& Integrity &  \\ 
		& Level &  \\
		& (SIL) & \\ \midrule
		Catastrophic & 4 & Loss of multiple lives\\
		Critical & 3 & Loss of a single life\\
		Marginal & 2 & Major injuries to one or more persons\\
		Negligible & 1 & Material damage, at most minor injuries\\
		No Consequence & 0 & No damages, except user dissatisfaction\\
		\bottomrule
	\end{tabular}
\end{table}

%Table 4: Approximate cross-domain mapping of safety levels based on data from [20]–[24].
\begin{table}\centering
	\caption{Approximate cross-domain mapping of safety levels based on data from~\cite{Verhulst2013, Verhulst2013a, Baufreton2010, Blanquart2012, Machrouh2012, Kafka2012}.} 
	\label{tab:sil-mapping}
	%\ra{1.3}
	\begin{tabular}{@{}ccccc@{}}\toprule
		Probability     & General           & Automotive
		& Aviation
		& Railway\\ 
		of Failure       & Programmable
		& ISO 26262
		& DO-178/254
		&CENELEC \\
		per Hour        & Electronics      &                     &                       & 50126 \\
		& IEC-61508       &                     &                       &  128/129\\
		\midrule
		-                                                                       & (SIL-0) & QM & DAL-E &  (SIL-0)\\
		10\textsuperscript{-6} – 10\textsuperscript{-5} & SIL-1 & ASIL-A & DAL-D & SIL-1 \\
		10\textsuperscript{-7} – 10\textsuperscript{-6} & SIL-2 & ASIL-B/C & DAL-C & SIL-2 \\
		10\textsuperscript{-8} – 10\textsuperscript{-7} & SIL-3 & ASIL-D & DAL-B & SIL-3 \\
		10\textsuperscript{-9} – 10\textsuperscript{-8} & SIL-4 & -          & DAL-A & SIL-4 \\
		\bottomrule
	\end{tabular}
\end{table}

To put probability of failure per hour of operation in context, we will compare it to safety standards across different industries. Table \ref{tab:sil-meaning} shows the typical breakdown of Safety Integrity Level (SIL) by hazard category. The strictest level (SIL-4) occurs where the consequence of failure is the loss of multiple human lives. The more lenient level, SIL-0, represent cases where the consequence of failure is only some dissatisfaction or discomfort. Table \ref{tab:sil-mapping} shows an approximate cross-domain mapping of aviation, rail, general programmable electronics, and automotive safety integrity levels. In rail, aviation, and programmable electronics, the strictest levels are those corresponding to failures which could cause the loss of multiple human lives and correspond to an integrity level of 10\textsuperscript{-9}~failures~/~hour. In rail and electronics this is SIL-4. In aviation, this is Design Assurance Level (DAL) A. The automotive industry's strictest requirement, Automotive Safety Integrity Level (ASIL) D is closer to SIL-3 and DAL-B in practice or 10\textsuperscript{-8}~--~10\textsuperscript{-7}~failures~/~hour~\cite{Kafka2012}. We're targeting 10\textsuperscript{-8}~failures~/~hour for the localization system, putting us in the range of ASIL-D. The virtual driver system will also be designed to ASIL-D standards, and hence it follows that subsystems like localization need to comply with this standard.

\section{Horizontal Requirements}
\label{Sec:Horizontal} 

\begin{figure}[h]
	\centering
	\subfloat[Straight road.]{\includegraphics[width=3.0in]{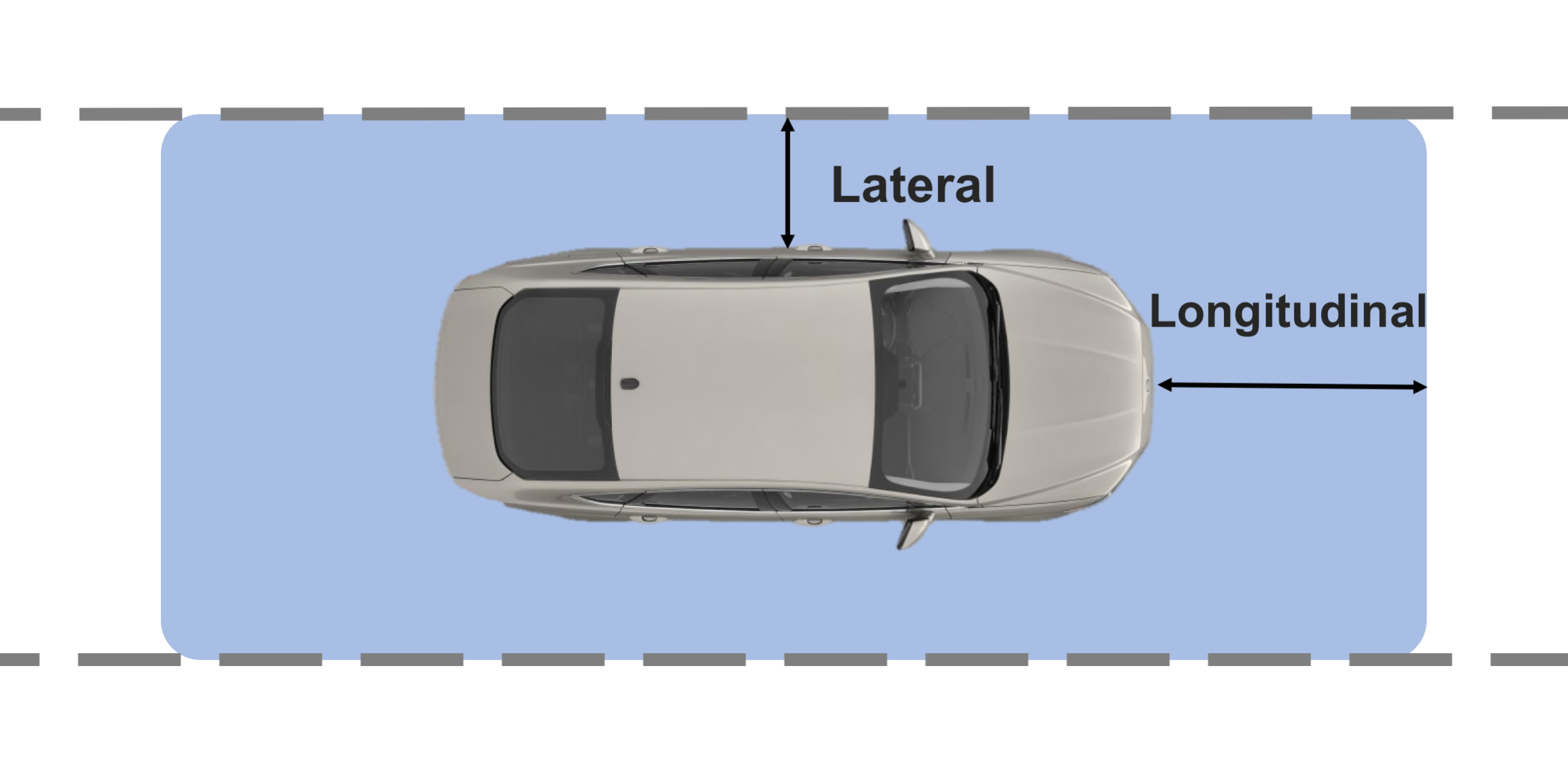}%
		\label{fig:req-straight}}
	\hfil
	\subfloat[Curved road.]{\includegraphics[width=3.0in]{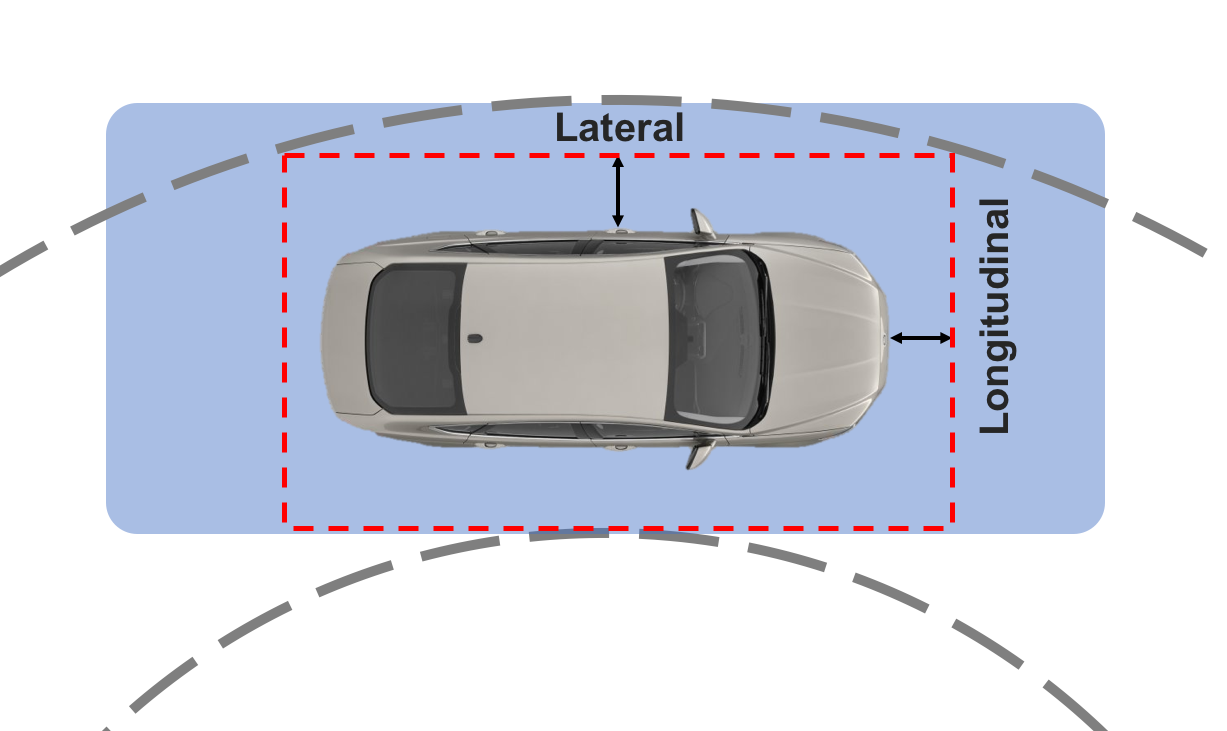}%
		\label{fig:req-curved}}
	\caption{Bounding box required for localization broken down by lateral and longitudinal components.}
	\label{fig:req-straight-vs-curved}
\end{figure}

The horizontal localization requirements for autonomous vehicles are a function of their physical dimensions and the road geometry. The goal is to keep the vehicle in its respective lane during typical operation. This leads to lateral and longitudinal localization requirements as shown in Figure \ref{fig:req-straight}. To scale, this shows the lateral clearance that can be expected with a mid-size sedan (e.g. a Ford Fusion) on a straight stretch of US freeway. This makes it appear as though lateral and longitudinal requirements are decoupled, but this is not entirely the case. Figure \ref{fig:req-curved} shows the coupling between these directions in turns, hence road curvature causes coupling between requirements in lateral and longitudinal directions. The analysis presented here will be focused on standards within the United States where assumptions will be made about typical vehicle width $w_v$, vehicle length $l_v$, road width $w$, and road radius of curvature $r$. A similar analysis could be undertaken with road and vehicle standards of other regions. 

%Figure 8: Bounding box required for localization on a straight road broken down by lateral and longitudinal components.

%Table 5: Vehicle dimension standards in the US [25], [26].
\begin{table}[h]\centering
	\caption{Vehicle dimension standards in the US~\cite{AmericanAssociationofStateHighwayandTransportationOfficials2001, U.S.DepartmentofTransportationFederalHighwayAdministration2017}. }
	\label{tab:vehicle-standards}
	\ra{1.3}
	\begin{tabular}{lccc}\toprule
		Vehicle Type	&	Width [m]	&	Length [m]	&	Height [m] \\ \midrule
		Passenger (P)	&	2.1	&	5.8	&	1.3
\\
		Single Unit Truck (SU)	&	2.4		&	9.2	&	3.4 - 4.1
\\
		City Bus	&	2.6	&	12.2	&	3.2
\\
		Semitrailer	&	2.4 - 2.6	&	13.9 - 22.4	&	4.1
\\
		\bottomrule
	\end{tabular}
\end{table}

We will begin with vehicle dimensions. Standards for road vehicle dimensions in the US are summarized in Table \ref{tab:vehicle-standards} and reflect maximum dimensions for different vehicle classes. A more detailed breakdown for some example passenger (P) vehicles is given in Table \ref{tab:vehicle-dim-typ}. This ranges from the subcompact to large 6-wheel `dualie' pickup trucks though the latter technically falls into the single unit truck (SU) category. As will be discussed, not all vehicles are meant for all roads, and hence some care must be taken when developing the localization requirements for vehicles. For example, semi-trucks are not meant to be driven on residential streets and hence requirements should not be set to roadways that are impossible for such a vehicle to navigate. Here, we will focus on passenger vehicles. 

%Table 6: Typical vehicle dimensions.
\begin{table}\centering
	\caption{Typical vehicle dimensions. }
	\label{tab:vehicle-dim-typ}
	\ra{1.3}
	\begin{tabular}{llccc}\toprule
		Vehicle Type	&	Example\textsuperscript{*} &	Width [m]	&	Length [m]	&	Height [m] \\ \midrule
		Subcompact & Fiesta	&	1.72	&	4.06	&	1.48 \\
		Compact	& Focus	&	1.82	&	4.54	&	1.47
\\
		Mid-Size	&	Fusion	&	1.85	&	4.87	&	1.48
\\
		Full-Size	&	Taurus	&	1.94	&	5.15	&	1.54
\\
		Crossover	&	Escape	&	1.84	&	4.52	&	1.68
\\
		Small SUV	& Edge	&	1.93	&	4.78	&	1.74
\\
		Standard SUV	&	Explorer	&	2.00	&	5.04	&	1.78
\\
		Van	& Transit	&	2.07-2.13\textsuperscript{**}	&	5.59-6.70\textsuperscript{**} &	2.09-2.76\textsuperscript{**}\\
		Pickup Truck	&	F-series	&	2.03-2.43\textsuperscript{**}		&	5.32-6.76\textsuperscript{**}	&	2.06
\\
		\bottomrule
	\end{tabular}
	\raggedright	
	\\	
	\rule{0pt}{2ex}  
	 *Based on 2018 Ford model year. \\
	**The wider F-series trucks \& Transits are dual wheeled or `dualies.'
\end{table}

Next is road geometry. Road curvature is a function of design speed and is based on limiting values of side friction factor $f$ and superelevation $e$~\cite{AmericanAssociationofStateHighwayandTransportationOfficials2001}. Superelevation is the rotation of the pavement on the approach to and throughout a horizontal curve and is intended to help the driver by countering the lateral acceleration produced by tracking the curve. The other important factor is road width, which typically ranges from 3.6~meters on standard freeways to 2.7~meters on limited residential streets~\cite{AmericanAssociationofStateHighwayandTransportationOfficials2001}. Road width and curvature are the elements that define the localization requirements to ensure the vehicle knows it is within its lane to the certainty level defined in Section \ref{Sec:Integrity}. The limiting cases for each road type have been assembled in Table \ref{tab:limiting-road-designs} for passenger type (P) vehicles. 

%Table 8: Minimum radius for the design of US roads using limiting values of rate of roadway superelevation e and coefficient of friction f [25]. 
% We are going to skip this since it is look-up able. 

%Table 9: Some limiting road design elements, based on designs for passenger (P) vehicles as defined in Table 5. Based on data from [25], [27], [28].
\begin{table}\centering
	\caption{Some limiting road design elements, based on designs for passenger (P) vehicles. Based on data from~\cite{AmericanAssociationofStateHighwayandTransportationOfficials2001, UnitedStatesDepartmentofTransportation2000, WashingtonStateDepartmentofTransportation2017}.}
	\label{tab:limiting-road-designs}
	\ra{1.5}
	\begin{tabular}{>{\raggedright\arraybackslash}m{0.9in}>{\centering\arraybackslash}m{0.6in}>{\centering\arraybackslash}m{0.6in}>{\centering\arraybackslash}m{0.6in}}\toprule
		Road Type	&	Design Speed [km/h]	&	Lane Width
[m]	&	Minimum Radius [m]
\\
		\midrule
		Freeway	&	80 - 130	& 	3.6	&	195\textsuperscript{**}
\\
		Interchanges	&	30 - 110	& 	3.6 - 5.4	&	150 - 15 
\\
		Arterial	&	50 - 100	& 	3.3 - 3.6	& 	70\textsuperscript{**} \\
		Collector	&	50	&	3.0 - 3.6	& 	70\textsuperscript{**} \\
		Local	&	20 - 50	&	2.7\textsuperscript{*} - 3.6	& 	10\textsuperscript{**} \\
		Hairpin~Turn / 
Cul-de-Sac	&	$<$ 20	&	6.0		&	7
\\
		Single Lane Roundabout	&	$<$ 20	&	4.3	&	11
\\
		\bottomrule
	\end{tabular}
	\raggedright
	\\ 			
	\rule{0pt}{2ex}  
	\hangindent=0.25cm
 *The lower bound of 2.7~m is the exception, not the rule, and is typically reserved for residential streets with low traffic volumes. 
\\
	\hangindent=0.25cm
**Based on design speeds and limiting values of rate of roadway superelevation $e$ and coefficient of friction $f$. 
\end{table}

 %Figure 10: Bounding box geometry in a turn. This shows the allowable maximum position error of the vehicle to ensure it is within the lane known as the alert limits. 
\begin{figure}
	\centering
	{\includegraphics[width=3.3in]{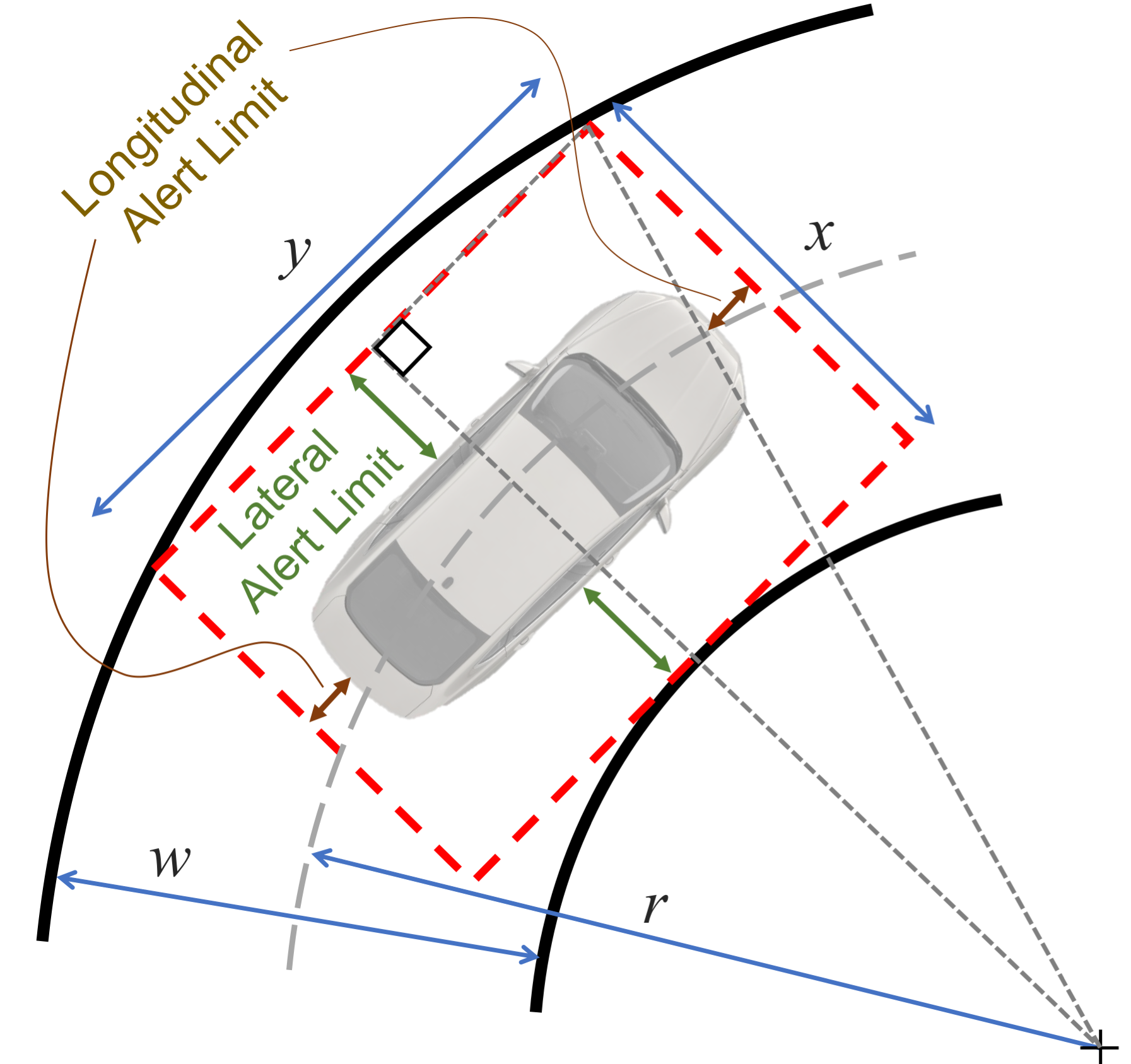}
		\caption{Bounding box geometry in a turn. This shows the allowable maximum position error of the vehicle to ensure it is within the lane known as the alert limits. }
		\label{fig:alert-limit-bnd-box}}
\end{figure}

The relationship between the road width and curvature and the interior bounding box around the vehicle is shown in Figure \ref{fig:alert-limit-bnd-box}. The relationship between the lateral and longitudinal bounds is found by Pythagoras: 
\begin{equation} \label{eq:pyth1}
\left( \frac{y}{2}\right)^2 + \left( r - \frac{w}{2} + x \right)^2 = \left( r + \frac{w}{2} \right)^2
\end{equation}
Solving for $x$ results in: 
\begin{equation} \label{eq:pyth2}
x = \sqrt{   \left( r + \frac{w}{2} \right)^2  - \left( \frac{y}{2}\right)^2} + \frac{w}{2} - r
\end{equation}
This allows us to determine the dimensions of the bounding box given the road geometry. In turn, given the vehicle dimensions, corresponding maximum permissible lateral and longitudinal errors (alert limits) can be derived. Lateral and longitudinal alert limits are a trade-off and there is a certain budget between them which is dependent on the road type. The lateral localization requirements are coupled to longitudinal through the allowed curvature and width of the road. For example, on the highway at high speed, the allowable road curvature is minimal and roads are fairly straight. This allows for the bounding box length to be large longitudinally before lateral requirements are overly constrained. Ultimately, choosing length $y$ fixes width $x$ or vice versa, and the resulting lateral and longitudinal alert limits are related to the vehicle length $l_v$ and width $w_v$ as follows: 
 \begin{equation} \label{eq:lat-alert-lim}
 \begin{split}
 \text{Lateral Alert Limit} &= \left( x-w_v \right) / 2 \\
  \text{Longitudinal Alert Limit} &= \left( y-l_v \right) / 2
 \end{split}
 \end{equation}

\begin{figure}
	\centering
	{\includegraphics[width=3.5in]{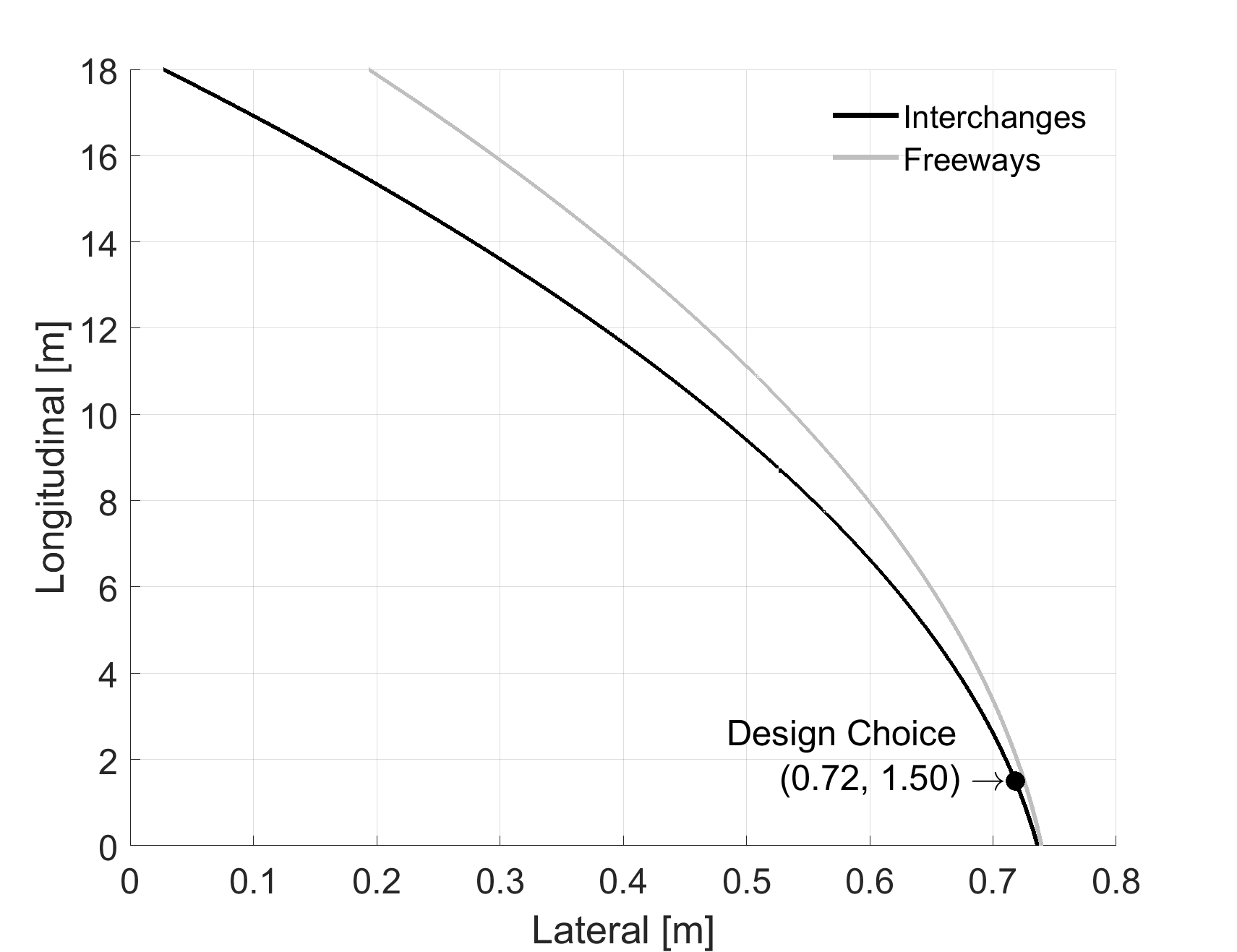}
		\caption{Lateral and longitudinal alert limit trade off for freeway and interchange geometry and passenger vehicle dimension limits. This is limited by lane widths of 3.6 meters with a minimum curvature of 150 meters.}
		\label{fig:freeway-al}}
\end{figure}

Using equations (\ref{eq:pyth2}) and (\ref{eq:lat-alert-lim}), the trade-off between lateral and longitudinal alert limits for freeways assuming  passenger vehicle design limits is shown in Figure \ref{fig:freeway-al}. This shows that as the longitudinal requirements are relaxed to several meters, the lateral requirements become more stringent. However, ultimately on/off ramps must be found within a reasonable tolerance on the freeway, so there is a more stringent longitudinal requirement based on vehicle operation. This is also constrained by situations where vehicles may be operating collaboratively and sharing their location via communications channels (V2X)~\cite{Feng2018}. In this design study, we limit the longitudinal alert limit to be less than half the length of a subcompact vehicle or 1.5~meters, well within the limits of reasonable highway following distances (even with the combined errors of two vehicles operating collaboratively) and the vehicle's ability to appropriately find on/off-ramps. With this longitudinal design number, we use Figure \ref{fig:freeway-al} to determine the required lateral alert limit to be 0.72~meters. The lateral alert limit for other vehicle types is summarized in Table \ref{tab:loc-alert-limits}.

%Figure 11: Lateral and longitudinal alert limit budget for freeway and interchange geometry and passenger vehicle size design limits. This is limited by lane widths of 3.6 meters with a minimum curvature of 150 meters.

%Figure 12: Lateral and longitudinal alert limit budget for local road geometry and passenger vehicle size design limits. Narrow streets are assumed to be 3.0 meters wide with a minimum curvature of 20 meters or 3.3 meters wide with minimum curvature of 10 meters. Single lane roundabouts and hairpin / cul-de-sac geometry is included for comparison.
\begin{figure}
	\centering
	{\includegraphics[width=3.5in]{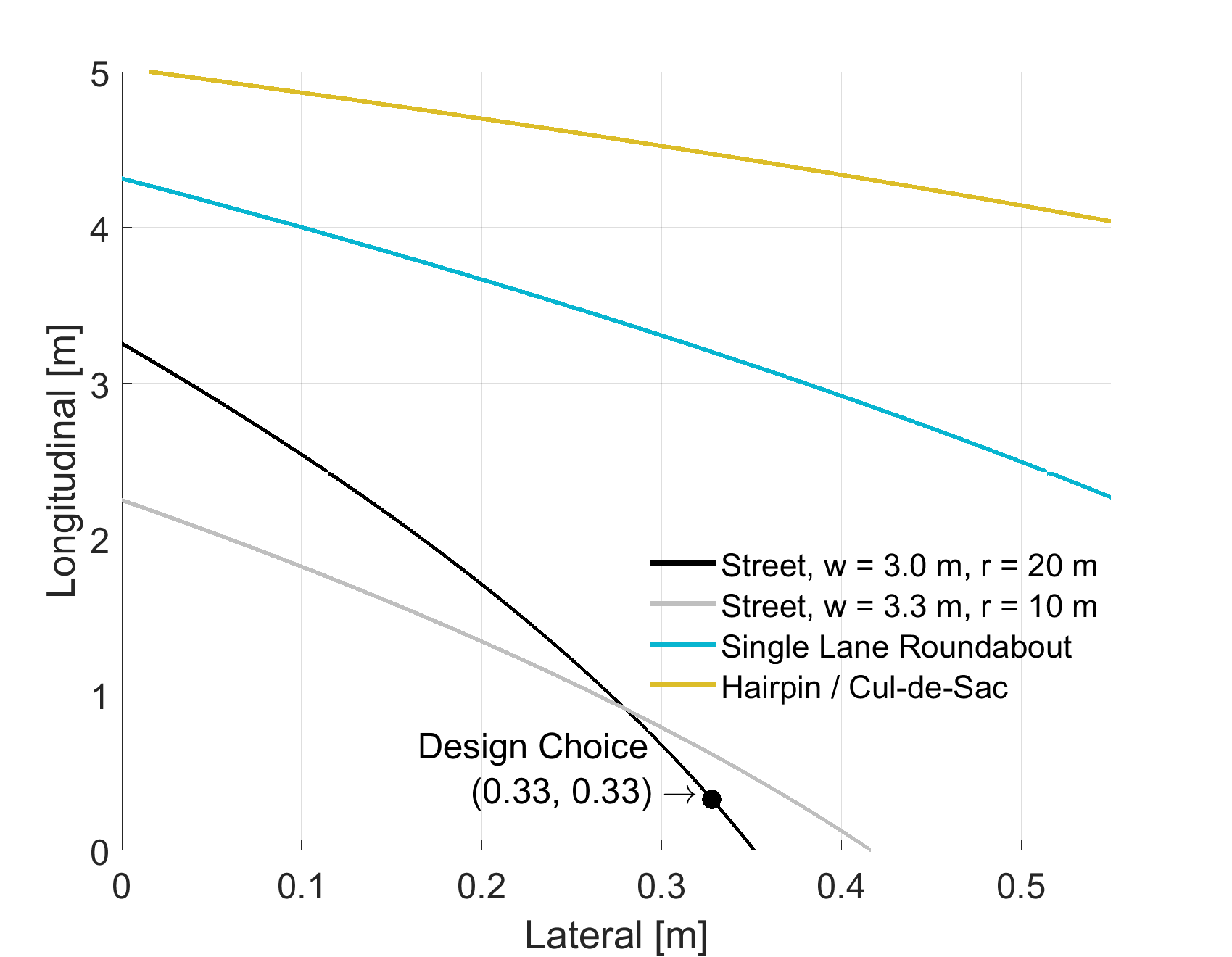}
		\caption{Lateral and longitudinal alert limit trade off for local road geometry and passenger vehicle dimension limits. Narrow streets are assumed to be 3.0~meters wide with a minimum curvature of 20 meters or 3.3~meters wide with minimum curvature of 10~meters. Single lane roundabouts and hairpin / cul-de-sac geometry is included for comparison.}
		\label{fig:local-al}}
\end{figure}

On the highway, lateral dominates the requirements since the coupling with curvature is negligible and is approximately 1 centimeter over the length of the largest pickup trucks. Hence, on the highway, the longitudinal alert limit is to some extent a design parameter. On local streets, with sharp turns, the curvature coupling results in tighter requirements in both directions. Figure \ref{fig:local-al} shows the trade-off between lateral and longitudinal alert limits for local road geometry for passenger vehicle limits. In this plot, we restrict the analysis to roads 3.3~meters wide with a curvature of 10~meters and 3.0~meters wide with a curvature of 20~meters. Also shown are the results for single lane roundabouts and hairpin turns. Though Table \ref{tab:limiting-road-designs} shows that some roads can be as narrow as 2.7~meters, this is the exception not the rule and we felt it too restricting to limit requirements based on this number. In addition, roads with tight curvature usually have wider lanes to accommodate as shown by the design recommendations for single lane roundabouts and hairpin turns / cul-de-sacs. Hence, these requirements are still conservative when neglecting 2.7~meters wide lanes~\cite{WashingtonStateDepartmentofTransportation2017}. 

For local streets, Figure \ref{fig:local-al} shows the trade-off between lateral and longitudinal alert limits. Thinking of limiting cases where vehicles are negotiating 90~degree turns, it seems logical that both directions become equally important to properly complete the maneuver, so the alert limits should be balanced equally in both directions. Figure \ref{fig:local-al} shows the equality point to be 0.33~meters for both the lateral and longitudinal alert limits for the largest passenger vehicles. Other vehicle types are summarized in Table \ref{tab:loc-alert-limits}. For scale, when operating in urban environments, 0.33~meters is also the minimum width of stop lines which are mandated to be between 0.3~and~0.6~meters (12~-~24~inches)~\cite{UnitedStatesDepartmentofTransportationFederalHighwayAdministration2009}.

\section{Vertical Requirements}
\label{Sec:Vertical} 

\begin{figure}
	\centering
	{\includegraphics[width=3.5in]{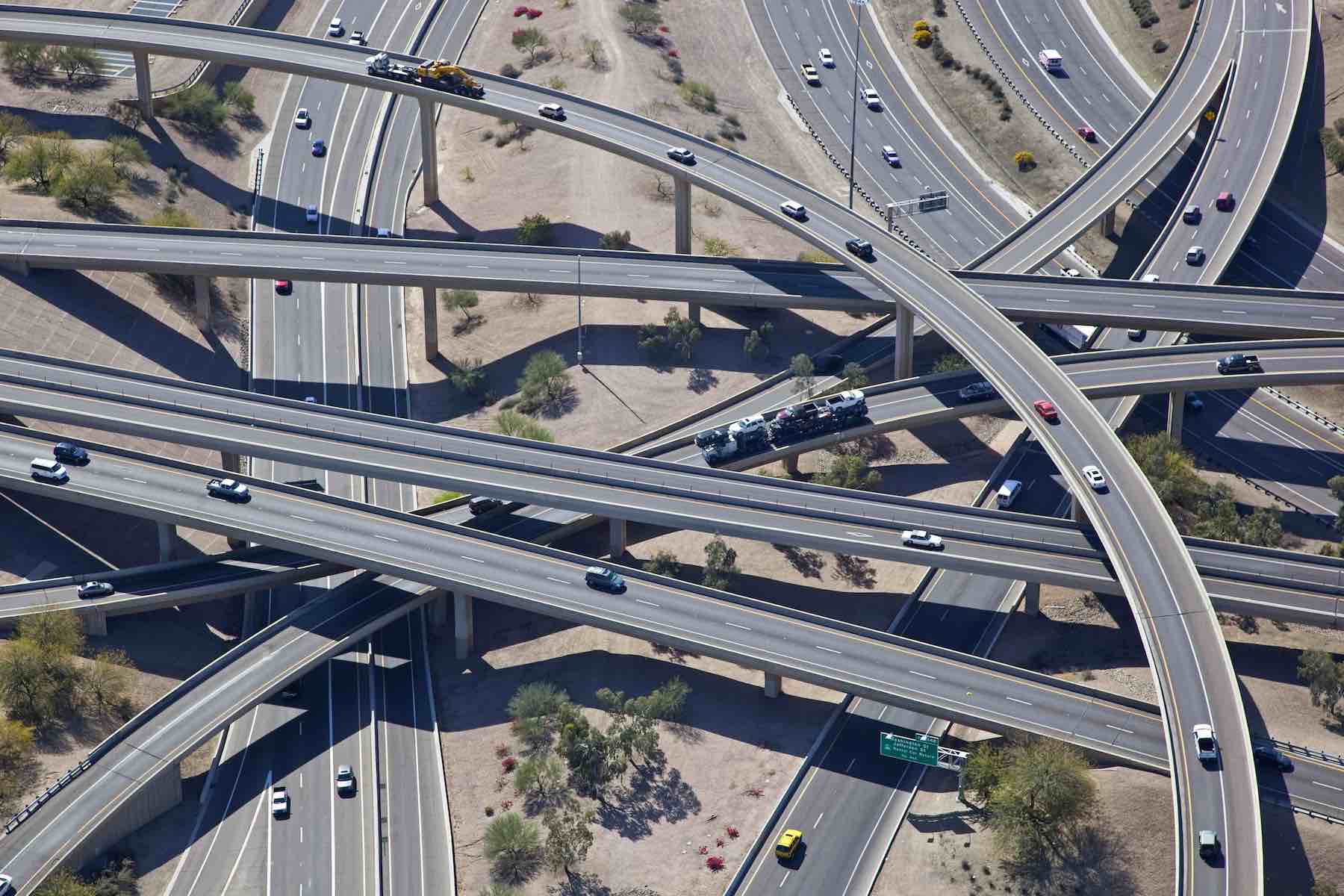}
		\caption{Example of a multi-level interchange in Phoenix, AZ. The `Mini Stack' is at the intersection of Interstate 10, State Route 51, and Loop 202.}
		\label{fig:high-five}}
\end{figure}

The recommended minimum vertical clearance for roads in the US is 4.4~meters (14.5~feet) [25]. This standard drives the permissible vertical height, including load, to be between 4.1~meters (13.5~feet) and 4.3~meters (14.5~feet), though this varies somewhat by state~\cite{AmericanAssociationofStateHighwayandTransportationOfficials2001, U.S.DepartmentofTransportationFederalHighwayAdministration2017}. Hence to reliably determine which road level we are on of an interchange for example, we must know our position to a fraction of this clearance height. Bounding our vertical position to $\pm$ half of this clearance is insufficient since this leaves a potential ambiguity on multi-decked roads or interchanges. An example of such an interchange is the `Mini Stack' in Phoenix, Arizona shown in Figure \ref{fig:high-five}. This multi-level interchange is at the intersection of Interstate 10, State Route 51, and Loop 202. For certainty, one-third of the minimum vertical clearance should be sufficient to resolve the ambiguity of which road level the vehicle is on. Hence the required Vertical Alert Limit (VAL) is: 
  \begin{equation} \label{eq:val}
\text{VAL} = \frac{ \text{min. vertical clearance}}{3}  =  \frac{4.4 ~\text{m}}{3} = 1.47 ~\text{m}
\end{equation}
Here, the vertical alert limit is vehicle independent (i.e. not dependent on vehicle dimensions) since it is used only to determine the road level. This differs from horizontal requirements developed in Section \ref{Sec:Horizontal} which strive to maintain a vehicle of certain dimensions within the bounds of the lane. This is reflected in Table \ref{tab:loc-alert-limits} which summarizes the lateral, longitudinal, and vertical alert limits for several vehicle types and different road operations including freeways / interchanges and local roads. 

It should be noted that this vertical positioning analysis has some limitations. As this analysis is seen as the far reaching goal for highly automated systems, the assumption here is that the vehicle will have a form of map to help in resolving position. In the interim, other applications such as V2X will likely not have maps. In the V2X scenario, the limiting factor for the vertical requirement is the trajectory estimation to determine whether or not a principal other vehicle is on a collision path with the subject vehicle. That is, elevation error will make a grade-separate interaction appear to be an at-grade crossing with collision potential. This is a more complex analysis to perform which is outside the scope of this paper.

%Table 10: Horizontal (lateral / longitudinal) and vertical localization alert limit requirements for US freeways and local roads.
\begin{table}\centering
	\caption{Horizontal (lateral / longitudinal) and vertical localization alert limit requirements for US freeways and local roads.}
	\label{tab:loc-alert-limits}
	\ra{1.6}
	\begin{tabular}{>{\arraybackslash}m{0.8in}|>{\centering\arraybackslash}m{0.25in}>{\centering\arraybackslash}m{0.25in}>{\centering\arraybackslash}m{0.25in}|>{\centering\arraybackslash}m{0.25in}>{\centering\arraybackslash}m{0.25in}>{\centering\arraybackslash}m{0.25in}}
		\toprule
		\multirow{2}{*}	{Vehicle Type} & \multicolumn{3}{c|}{Local Roads} & \multicolumn{3}{c}{Freeways \& Interchanges} \\ 			
		& Lat. [m] & Long. [m] & Vert. [m] &	Lat. [m] & Long. [m] & Vert. [m]  \\
		\midrule
		Mid-Size	&	0.48	&	0.48	&	1.47	&	0.85	&	1.50	&	1.47\\
		Full-Size	&	0.42	&	0.42	&	1.47	&	0.80	&	1.50	&	1.47\\
		Standard Pickup	&	0.38	&	0.38	&	1.47	&	0.76	&	1.50	&	1.47\\
		Passenger Vehicle Limits	&	0.33	&	0.33	&	1.47	&	0.72	&	1.50	&	1.47\\
		6-Wheel Pickup	&	-	&	-	&	-	&	0.56	&	1.50	&	1.47\\
		\bottomrule
	\end{tabular}
\end{table}

\section{Orientation Requirements}
\label{Sec:Orientation}

\begin{figure}[h]
	\centering
	{\includegraphics[width=3.4in]{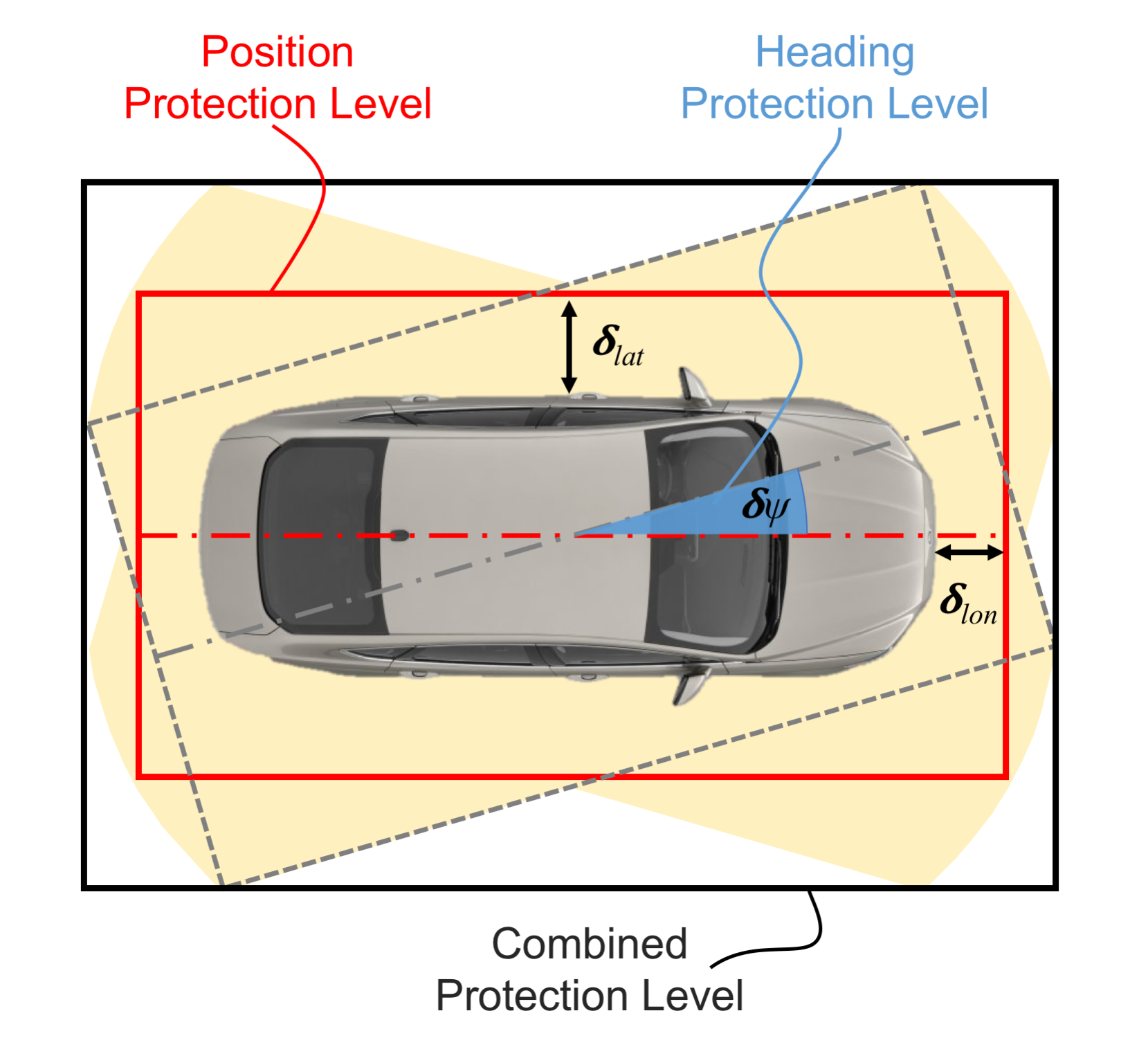}
		\caption{Combined effect of lateral / longitudinal position and heading errors on overall protection level. Heading errors rotate position errors and lead to a larger effective area of uncertainty.}
		\label{fig:combined-PL}}
\end{figure}

The horizontal and vertical alert limits discussed so far are the acceptable limit for all combined sources of error. As will be discussed in this section, this will include errors in both positioning and attitude (orientation). The vehicle attitude is described in terms of its roll $\theta$, pitch $\phi$, and heading $\psi$ angles. Errors in these parameters will rotate the position protection level box around the vehicle and result in a larger effective protection level. This effect is shown in Figure \ref{fig:combined-PL}. This shows how errors in heading and position map to a larger combined protection level area and hence why knowledge of attitude error is important. 

%Figure 13: Combined effect of lateral / longitudinal position and heading errors on overall position error. Heading errors rotate position errors and lead to a larger effective area of uncertainty. 

This effect leads to requirements on acceptable errors in roll, pitch, and heading as well as position. We will begin with mapping errors in position and orientation into a combined protection level. Assuming a box around the vehicle of width $x$, length $y$, and height $z$ (following the notation of Figure \ref{fig:alert-limit-bnd-box}), maximum errors in roll $\pm \delta \theta$, pitch $\pm \delta \phi$, and heading $\pm \delta \psi$ angles are mapped through the following Euler sequence: 
\begin{equation} \label{eq:combined-errors-full}
\mathbf{x}'=\mathbf{R}_3\left( \pm \delta \psi \right) \mathbf{R}_2\left( \pm \delta \theta \right) \mathbf{R}_3\left( \pm \delta \phi \right) \mathbf{x}
\end{equation}
where $\mathbf{x}$ is the dimensions of the box $[x, y, z]^T$ reflecting position protection level, $\mathbf{x}'$ is the dimensions of the inflated box $[x', y', z']^T$ representing the protection level from both positioning and orientation, and $\mathbf{R}_i$ are the following rotation matrices:  

\begin{equation} \label{eq:R1}
\mathbf{R}_{1}\left( \pm \delta \phi\right)  = 
\begin{bmatrix}
1 & 0 & 0 \\
0 & \cos(\pm \delta \phi) & -\sin(\pm \delta \phi) \\
0 & \sin(\pm \delta \phi) &  \cos(\pm \delta \phi)
\end{bmatrix}
\end{equation}

\begin{equation} \label{eq:R2}
\mathbf{R}_{2}\left( \pm \delta \theta\right)  = 
\begin{bmatrix}
\cos(\pm \delta \theta) & 0 & \sin(\pm \delta \theta) \\
0 & 1 & 0 \\
-\sin(\pm \delta \theta) & 0 & \cos(\pm \delta \theta)
\end{bmatrix}
\end{equation}

\begin{equation} \label{eq:R3}
\mathbf{R}_{3}\left(\pm \delta  \psi\right)  = 
\begin{bmatrix}
\cos(\pm \delta \psi) & -\sin(\pm \delta \psi) & 0\\
\sin(\pm \delta \psi) & \cos(\pm \delta \psi) & 0\\
0 & 0 & 1 \\
\end{bmatrix}
\end{equation}
The position protection level $\mathbf{x}$ is related to errors in lateral $\delta_{lat}$, longitudinal $\delta_{lon}$, and vertical $\delta_{vert}$ positioning as follow: 
%x=w_v+2δ_lat
%y=〖l_v+2δ〗_lon
%z=2 δ_vert	(12)
\begin{equation} \label{eq:position-pl}
\mathbf{x}  = 
\begin{bmatrix}
x \\ y \\ z
\end{bmatrix}
= \begin{bmatrix}
w_v + 2\delta_{lat} \\ l_v + 2\delta_{lon} \\ 2\delta_{vert}
\end{bmatrix}
\end{equation}

We are after the worst-case error bounds, which are obtained by letting all the terms constructively add by setting $\cos(\pm \delta \cdot) \rightarrow  \cos(\delta \cdot)$ and $\pm \sin(\pm \delta \cdot) \rightarrow \sin(\delta \cdot)$. By necessity, errors in orientation will also have to be small, meaning $\delta \theta$, $\delta \phi$, and $\delta \psi$ will be $\ll$~1~radian (57 degrees). This allows us to make a small angle approximation to simplify these equations, where $\cos(\delta \cdot) \rightarrow 1$ and $\sin(\delta \cdot) \rightarrow \delta \cdot$. Multiplying out and neglecting higher order terms results in the following: 

\begin{equation} \label{eq:linearized-matrix-form}
\mathbf{x}'  = 
\begin{bmatrix}
1 & \delta \psi & \delta \theta \\
\delta \psi & 1 & \delta \phi \\
\delta \theta & \delta \phi & 1 \\
\end{bmatrix}
\mathbf{x}
\end{equation}
Combining equations (\ref{eq:position-pl}) and (\ref{eq:linearized-matrix-form}) along with our definition of protection levels given by Figure \ref{fig:PL-def} (where now Lat.~PL~$=(x'-w_v)/2$, Lon.~PL~$=(y'-l_v)/2$, and VPL~$=z'/2$) gives the combined protection level as a function of position and orientation errors: 

\begin{equation} \label{eq:linearized-combined-pl}
\begin{split}
\text{Lat. PL} &= \delta_{lat} + \left( \delta_{lon} + l_v/2\right) \delta \psi + \delta_{vert} \delta \theta \\
\text{Lon. PL} &= \delta_{lon} + \left( \delta_{lat} + w_v/2\right)  \delta \psi + \delta_{vert} \delta \phi \\
\text{VPL} &= \delta_{vert} + \left( \delta_{lat} + w_v/2\right) \delta \theta + \left( \delta_{lon} + l_v/2\right) \delta \phi
\end{split}
\end{equation}

\begin{figure}[h]
 	\centering
 	{\includegraphics[width=3.4in]{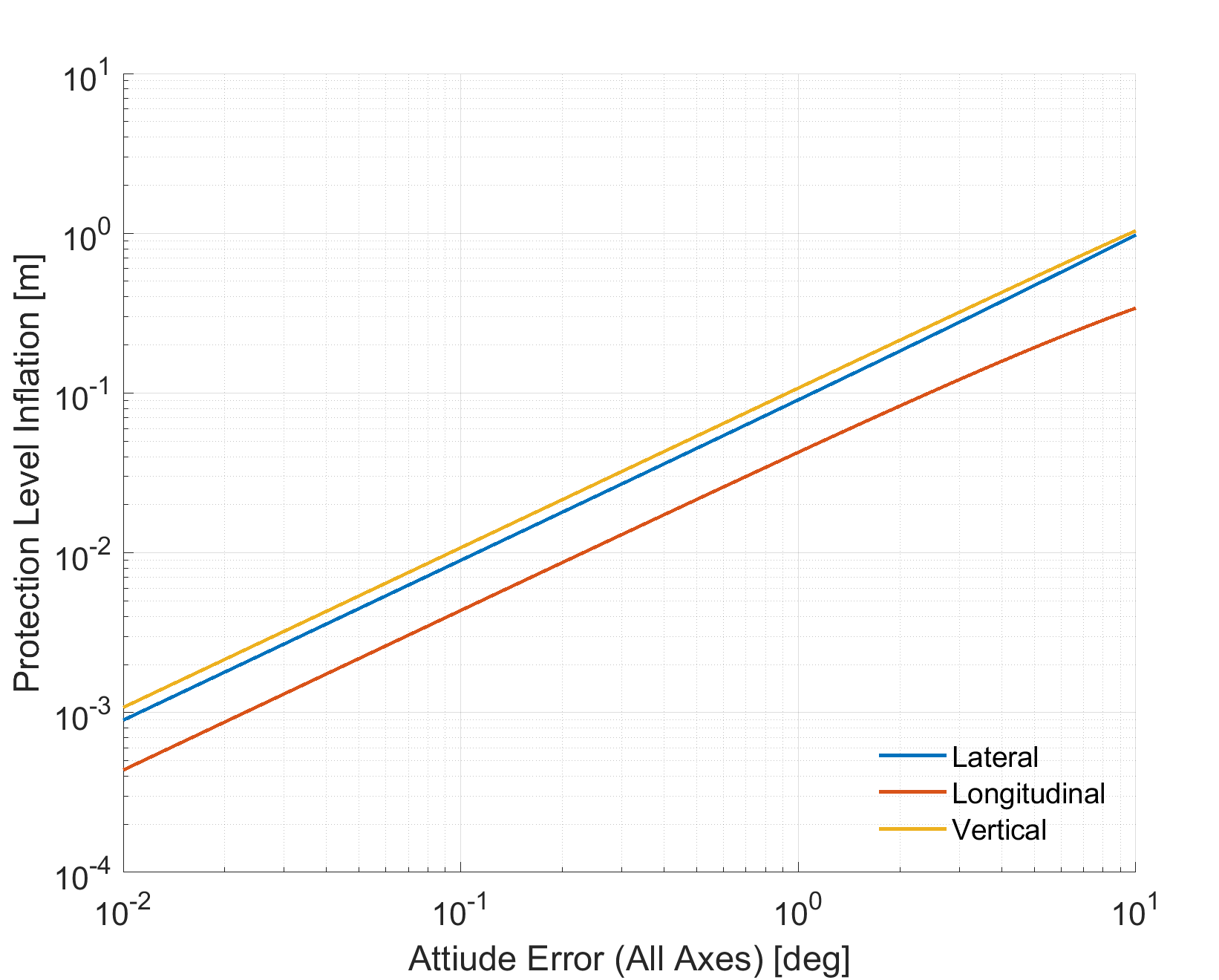}
 		\caption{Example of protection level inflation as a function of attitude error (on all axes). This assumes passenger vehicle design limits and the highway/interchange alert limits given in Table \ref{tab:loc-alert-limits}.}
 		\label{fig:attitude-PL-inflation}}
 \end{figure}

To give a sense of how the above equations scale, Figure \ref{fig:attitude-PL-inflation} shows the lateral, longitudinal, and vertical protection level inflation as a function of attitude error for the freeway alert limits given in Table \ref{tab:loc-alert-limits}. This assumes the same angular error in each direction and shows how quickly these inflate our protection levels. The allocation of position and orientation error budgets is ultimately a design choice which will be examined in more detail in Section \ref{Sec:Design}.  

%Figure 14: Example of protection level inflation as a function of attitude error (on all axes). This assumes passenger vehicle design limits and the highway/interchanges alert limits given in Table 9. 

\section{Update Frequency}
\label{Sec:Frequency} 

\begin{figure}[h]
	\centering
	{\includegraphics[width=3.5in]{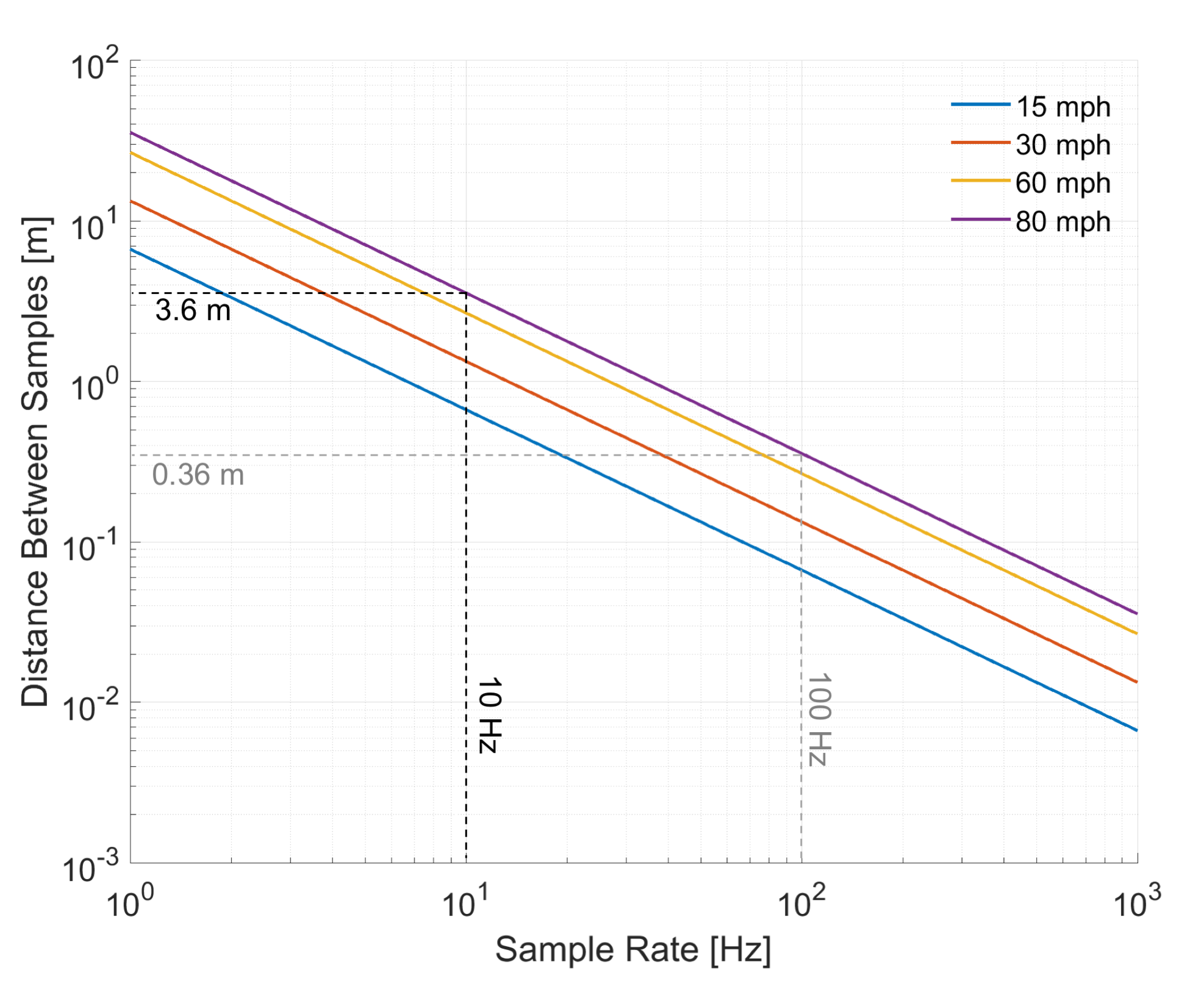}
		\caption{The relationship between sample rate, speed, and distance between samples. }
		\label{fig:freq-speed-dist}}
\end{figure}

The time required between successive localization updates (latency) is a function of the vehicle speed and road geometry. The longer the update interval, the larger the distance between localization updates. For example, at 100~km/h~(62~mph), 10~Hz gives localization updates 2.7~meters apart, the lane width of some local streets. At 130~km/h~(80~mph), 10~Hz yields 3.6~meters between successive updates, the width of a freeway lane. The relationship between vehicle speed, sampling rate, and the distance between samples is given in Figure \ref{fig:freq-speed-dist}. A lag in position update leads directly to further uncertainty in localization, predominantly in the longitudinal direction. Hence this lag must be managed such that it does not become a dominant factor. 

% Figure 15: The relationship between sample rate, speed, and distance between samples. 

Section \ref{Sec:Horizontal} showed that highway operation requires a longitudinal protection level of 1.5~meters. At highway speeds of up to 130~km/h~(80~mph), 100~Hz gives rise to 0.36~meter spacing between successive position updates and 200~Hz gives 0.17~meters. This drives our requirement since we want the contribution of this uncertainty to be only a small component of our protection level. An update of 200~Hz corresponds to a successive point spacing one tenth of our chosen alert limit and seems most appropriate. An update of 200~Hz may seem fast for localization technologies, where LiDAR and GNSS typically output position updates at 10~--~20~Hz, but when combined with an inertial measurement unit (IMU), rates of 200~Hz can be achieved. This requirement is ultimately that on the system as a whole, not each piece individually. 

The update rate can be throttled based on speed, slowing during low speed driving to save compute and power, and to increase range. For example, at 100~km/h~(62~mph) on the freeway, one-tenth the longitudinal alert limit can be achieved at 150~Hz. However, as was shown in Section \ref{Sec:Horizontal}, operation on local streets requires tighter requirements. Even on local streets where sharp turns are taken at 15~km/h~(10~mph), we require our alert limit to be 0.33~meters, one tenth of this number at this speed corresponds to 125~Hz. Hence, 100~Hz or greater appears to be the appropriate update rate for both highway and local street operation. 

\begin{comment}
This update frequency can be reduced with appropriate knowledge of the vehicle dynamic model and/or an accurate measure of velocity and angular rates. To give a feel for this, if we had a good measurement of longitudinal velocity while traveling in a straight line, the longitudinal position error $\delta_{lon}$ between successive updates is:  
\begin{equation} \label{eq:speed-error}
\delta_{lon} = x_{lon, t} - x_{lon, t-1} + v_{lon, t-1} \delta t
\end{equation}
where $x_{lon, t}$ and $x_{lon, t – 1}$ is our longitudinal position at time $t$ and $t – 1$, respectively, $v_{lon, t – 1}$ is the longitudinal speed at time $t – 1$ and $\delta t$ is the time between successive updates. With this, our position uncertainty doesn't drift nearly as much between successive updates and our error grows with our uncertainty in velocity and hence much more slowly. This is ultimately more complex and is a function of the vehicle dynamics along with uncertainties in velocity and angular rates, but it shows that the position update frequency can be reduced in certain conditions. 
\end{comment}

\section{Localization Requirements Design}
\label{Sec:Design} 

In this section we will summarize the design process for allocating localization requirements. This balances allowable errors in position and attitude. The results are summarized in Tables \ref{tab:loc-req-highway} and \ref{tab:loc-req-local}. 

Our design equations are based on (\ref{eq:linearized-combined-pl}), where we require protection~levels~$\leq$~alert~limits. This guarantees knowledge that we are within the desired lane and on the appropriate road level to the degree of safety needed for operation. These design equations are as follows: 

\begin{equation} \label{eq:design-all}
\begin{split}
\delta_{lat} + \left( \delta_{lon} + l_v/2\right) \delta \psi + \delta_{vert} \delta \theta &\leq \text{Lat. AL} \\
\delta_{lon} + \left( \delta_{lat} + w_v/2\right)  \delta \psi + \delta_{vert} \delta \phi &\leq \text{Lon. AL} \\
\delta_{vert} + \left( \delta_{lat} + w_v/2\right) \delta \theta + \left( \delta_{lon} + l_v/2\right) \delta \phi &\leq \text{VAL} 
\end{split}
\end{equation}

%\begin{equation} \label{eq:design-lon}
%\delta_{lon} + \left( \delta_{lat} + w_v/2\right)  \delta \psi + \delta_{vert} \delta \phi \leq \text{Lon. AL} 
%\end{equation}
%
%\begin{equation} \label{eq:design-vert}
%\delta_{vert} + \left( \delta_{lat} + w_v/2\right) \delta \theta + \left( \delta_{lon} + l_v/2\right) \delta \phi \leq \text{VAL} 
%\end{equation}

In the above, our protection levels are written as a function of both position and orientation errors as developed in Section \ref{Sec:Orientation}. These coupled equations must satisfy the constraints developed in Sections \ref{Sec:Horizontal} and \ref{Sec:Vertical}, which describe the total allowable lateral, longitudinal, and vertical errors (alert limits) as a function of the road geometry and vehicle dimensions. These alert limits are summarized in Table \ref{tab:loc-alert-limits}. 

We are most constrained in horizontal components, especially the lateral direction, so we will use this as our driving constraint equation. Assuming angular errors are allowed to be the same in each direction, namely $\delta \theta = \delta \phi = \delta \psi = \delta \lambda$, the lateral component of (\ref{eq:design-all}) simplifies to: 
\begin{equation} \label{eq:design-lat-simple}
\delta_{lat} + \left( \delta_{lon} +  \delta_{vert} + l_v/2\right) \delta \lambda \leq \text{Lat. AL}
\end{equation}
Section \ref{Sec:Horizontal} showed that for passenger vehicle limits, the sum of allowable longitudinal and vertical errors for freeway operation turns out to be approximately half the vehicle length $l_v/2$, so a good rule of thumb is: 
\begin{equation} \label{eq:design-lat-final}
\delta_{lat} + l_v \delta \lambda \leq \text{Lat. AL}
\end{equation}
Since the limiting $l_v$ for passenger vehicles is 5.8~meters and the lateral alert limit was set at 0.72~meters for freeway operation (see Table \ref{tab:loc-alert-limits}), the acceptable error in orientation $\delta \lambda$ must be less than 0.1~radians (5.73~degrees) otherwise we quickly exceed this limit. A reasonable choice for $\delta \lambda$ seems to be an orientation error of 1.5~degrees (0.03~radians) which leads to a contribution of 0.15~meters when scaled by $l_v$. This leads to a required lateral positioning error $\delta_{lat}$ limit of 0.57~meters to meet our combined requirement. 

\begin{table*}\centering
	\caption{Localization requirements for US freeway operation with interchanges. This assumes minimum lane widths of 3.6~meters and allowable speeds up to 137~km/h (85~mph).} 
	\label{tab:loc-req-highway}
	\ra{1.5}
	\begin{tabular}{>{\arraybackslash}m{0.8in}|>{\centering\arraybackslash}m{0.4in}>{\centering\arraybackslash}m{0.4in}>{\centering\arraybackslash}m{0.4in}>{\centering\arraybackslash}m{0.4in}|>{\centering\arraybackslash}m{0.4in}>{\centering\arraybackslash}m{0.4in}>{\centering\arraybackslash}m{0.4in}>{\centering\arraybackslash}m{0.4in}|>{\centering\arraybackslash}m{0.95in}}
		\toprule
		\multirow{2}{*}	{Vehicle Type} & \multicolumn{4}{c|}{Accuracy (95\%)} & \multicolumn{4}{c|}{Alert Limit} & \multirow{2}{0.75in} {\centering Prob. of Failure (Integrity)}  \\ 
		
		& Lateral [m] & Long. [m] & Vertical [m] & Attitude\textsuperscript{*} [deg] & Lateral [m] & Long. [m] & Vertical [m] & Attitude\textsuperscript{*} [deg]  & \\
		\midrule
		Mid-Size	& 0.24	&	0.48	&	0.44	&	0.51	&	0.72	&	1.40	&	1.30	&	1.50		&	10\textsuperscript{-9}~/~mile
		(10\textsuperscript{-8}~/~hour)\\
		Full-Size	&	0.23	&	0.48	&	0.44	&	0.51	&	0.66	&	1.40	&	1.30	&	1.50 &	10\textsuperscript{-9}~/~mile (10\textsuperscript{-8}~/~hour) \\
		Standard Pickup		&	0.21	&	0.48	&	0.44	&	0.51	&	0.62	&	1.40	&	1.30	&	1.50		&	10\textsuperscript{-9}~/~mile (10\textsuperscript{-8}~/~hour) \\
		Passenger Vehicle Limits	&	0.20	&	0.48	&	0.44	&	0.51	&	0.57	&	1.40	&	1.30	&	1.50	&	10\textsuperscript{-9}~/~mile (10\textsuperscript{-8}~/~hour) \\
		6-Wheel Pickup	&	0.14	&	0.48	&	0.44	&	0.51	&	0.40	&	1.40	&	1.30	&	1.50		&	10\textsuperscript{-9}~/~mile (10\textsuperscript{-8}~/~hour) \\
		\bottomrule
	\end{tabular}
	%\raggedright
	\\ 			
	\rule{0pt}{2ex}  
	*Error in each direction (roll, pitch, and heading).
\end{table*}

\begin{table*}\centering
	\caption{Localization requirements for US local roads. This assumes lanes 3.0~meters wide with a minimum curvature of 20~meters or 3.3~meters wide with minimum curvature of 10~meters.} 
	\label{tab:loc-req-local}
	\ra{1.5}
	\begin{tabular}{>{\arraybackslash}m{0.8in}|>{\centering\arraybackslash}m{0.4in}>{\centering\arraybackslash}m{0.4in}>{\centering\arraybackslash}m{0.4in}>{\centering\arraybackslash}m{0.4in}|>{\centering\arraybackslash}m{0.4in}>{\centering\arraybackslash}m{0.4in}>{\centering\arraybackslash}m{0.4in}>{\centering\arraybackslash}m{0.4in}|>{\centering\arraybackslash}m{0.95in}}
		\toprule
		\multirow{2}{*}	{Vehicle Type} & \multicolumn{4}{c|}{Accuracy (95\%)} & \multicolumn{4}{c|}{Alert Limit} & \multirow{2}{0.75in} {\centering Prob. of Failure (Integrity)} \\ 
		
		& Lateral [m] & Long. [m] & Vertical [m] & Attitude\textsuperscript{*} [deg] & Lateral [m] & Long. [m] & Vertical [m] & Attitude\textsuperscript{*} [deg]  & \\
		\midrule
		Mid-Size	&	0.15	&	0.15	&	0.48	&	0.17	&	0.44	&	0.44	&	1.40	&	0.50	&	10\textsuperscript{-9}~/~mile
		(10\textsuperscript{-8}~/~hour)\\
		Full-Size	&	0.13	&	0.13	&	0.48	&	0.17	&	0.38	&	0.38	&	1.40	&	0.50	 &	10\textsuperscript{-9}~/~mile (10\textsuperscript{-8}~/~hour) \\
		Standard Pickup		&	0.12	&	0.12	&	0.48	&	0.17	&	0.34	&	0.34	&	1.40	&	0.50	&	10\textsuperscript{-9}~/~mile (10\textsuperscript{-8}~/~hour) \\
		Passenger Vehicle Limits	&	0.10	&	0.10	&	0.48	&	0.17	&	0.29	&	0.29	&	1.40	&	0.50	&	10\textsuperscript{-9}~/~mile (10\textsuperscript{-8}~/~hour) \\
		\bottomrule
	\end{tabular}
	%\raggedright
	\\ 			
	\rule{0pt}{2ex}  
	*Error in each direction (roll, pitch, and heading).
\end{table*}

Local streets have more stringent requirements. Though longitudinal requirements are tighter, equation (\ref{eq:design-lat-final}) is still a reasonable approximation of how errors scale. Since the lateral alert limit in these conditions is 0.33~meters for passenger vehicle limits (see Table \ref{tab:loc-alert-limits}), we require nearly a threefold improvement compared to freeway design numbers. This puts us around 0.5~degrees of orientation error $\delta \lambda$ which leads to an error contribution of 0.05~meters when scaled by $l_v$. This leaves us with an allowable lateral position error $\delta_{lat}$ of 0.29~meters. 

\begin{figure}[h]
	\centering
	{\includegraphics[width=3.5in]{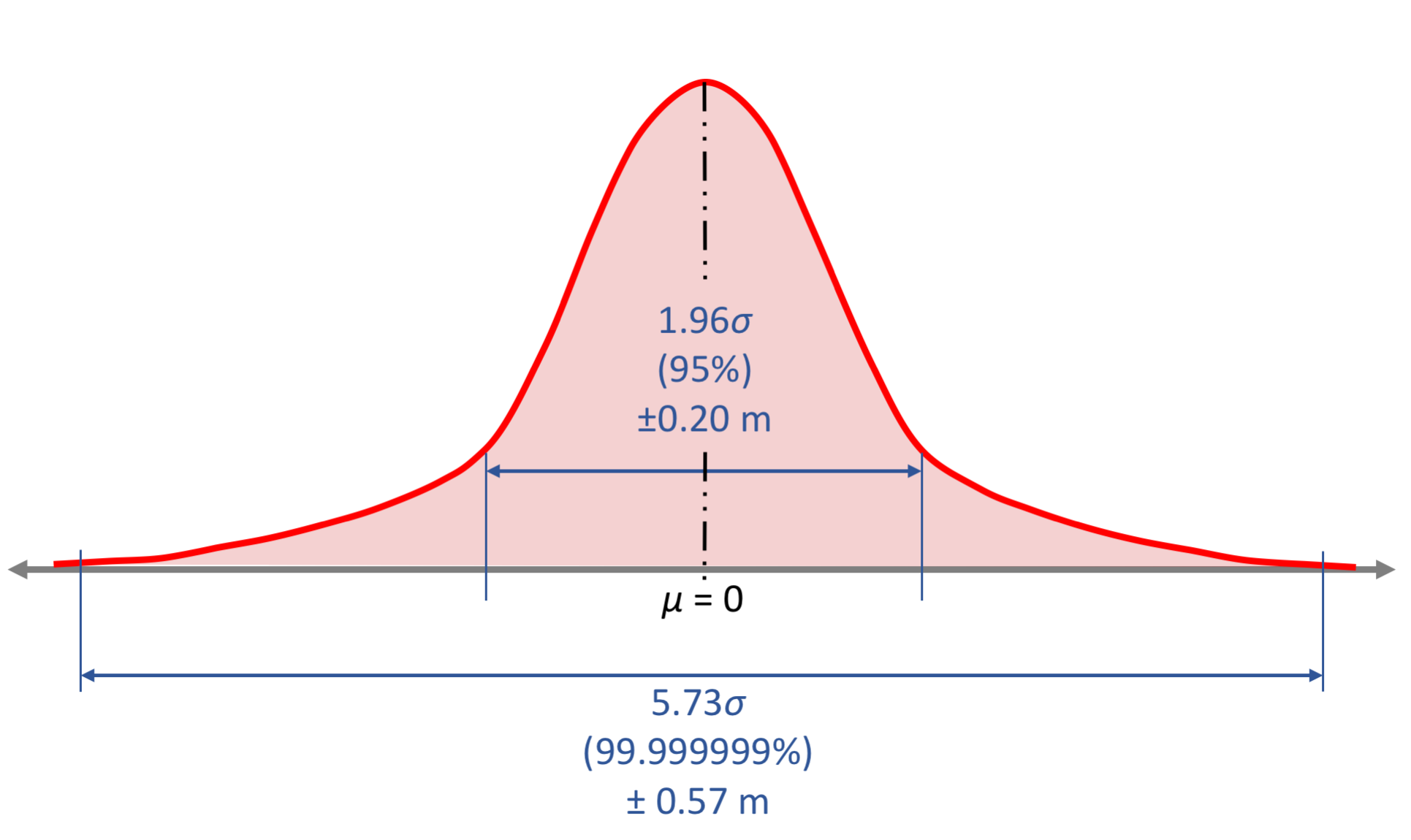}
		\caption{The desired error distribution for lateral positioning on freeways for passenger vehicle dimension limits, assuming a Gaussian distribution. This shows the 95\% accuracy at 0.20~meters and hard error bound at 0.57 m at 99.999999\% which is a probability of (1~--~10\textsuperscript{-8}).}
		\label{fig:error-dist}}
\end{figure}

Using design equations (\ref{eq:design-all}-\ref{eq:design-lat-final}) as a guide, along with the geometric bounds given in Table \ref{tab:loc-alert-limits} representing the total combined alert limits, bounds for position and orientation errors can be produced. To overload notation, we will also refer to these position and orientation bounds as alert limits. Using these numbers, we can obtain an approximation for the 95\% accuracy requirements by assuming a Gaussian distribution. Though the error distribution of the localization system may not be Gaussian, when thinking of this in Gaussian terms, (1~--~10\textsuperscript{-8}) is 99.999999\% or approximately 5.73$\sigma$. This gives us a sense when evaluating localization technologies of what statistics we should be looking for in terms of metrics like 95\% accuracy performance (1.96$\sigma$). In other words, when deriving hard error bounds (the alert limits) on localization requirements to a degree of certainty of (1~--~10\textsuperscript{-8}) we will take 95\% (1.96$\sigma$) accuracy as approximately one third of this number since 1.96$\sigma$~/~5.73$\sigma$ = 1~/~2.92. This relationship is shown visually in Figure \ref{fig:error-dist} for lateral freeway positioning requirements. Ultimately, there will be other considerations on the distribution of localization errors including smoothness of output and additional parameters such as acceleration and jerk which are relevant to controlling the vehicle for passenger comfort~\cite{Smith1978, OBrien1996}.

%Figure 7: The desired error distribution for lateral positioning on freeways for passenger vehicle dimension limits. This shows the 95% accuracy at 0.20 meters and hard error bound at 0.57 m at 99.999999% which is a probability of 1-10-8 as will be summarized in Table 11.

Putting all of the information developed so far together, requirements can be broken down by road type and operation. The requirements for freeways and interchanges are summarized in Table \ref{tab:loc-req-highway} for a variety of vehicles ranging from mid-size to large `dualie' pickup trucks. This includes position and attitude alert limits, 95\% accuracy, and the integrity requirements developed in Section \ref{Sec:Integrity}. The requirements for local roads are summarized in Table \ref{tab:loc-req-local} for passenger vehicles. Though speeds are lower, the road geometry is tighter, leading to more stringent requirements on localization. 

These results indicate that highway operations will require lateral accuracies in the 0.2~meters~(95\%) range, a conclusion which matches the requirements for lane departure warning systems \cite{Feng2018, EDMapConsortium2004}. Longitudinal and vertical requirements are more forgiving, with numbers in the 0.4~meters~(95\%) range, with pointing requirements in each direction of 0.5~degrees (0.01~radians) (95\%). Operations on local roads require lateral and longitudinal accuracies in the 0.1~meters~(95\%) range with pointing requirements of 0.17~degrees (3~milliradians) (95\%).

\section{Conclusion}
\label{Sec:Conclusion} 

The localization requirements for autonomous vehicles represent the next order of magnitude in accuracy needs for widespread deployment. Here, localization requirements in terms of accuracy, integrity, and latency were developed based on vehicle dimensions, road geometry standards, and a target level of safety. Integrity risk allocation leveraged the approach taken in civil aviation where similar requirements on localization are derived at 10\textsuperscript{-8}~probability of failure per hour of operation. Combining this with road geometry standards, requirements emerge for different road types and operation. For passenger vehicles operating on freeways, the result is a required lateral error bound of 0.57~m (0.20~m,~95\%), a longitudinal bound of 1.40~m (0.48~m,~95\%), a vertical bound of 1.30~m (0.43~m,~95\%), and an attitude bound in each direction of 1.50~deg (0.51~deg,~95\%). On local streets, the road geometry makes requirements more stringent where lateral and longitudinal error bounds of 0.29~m (0.10~m,~95\%) are needed with an orientation requirement of 0.50~deg (0.17~deg,~95\%).

It should be emphasized that these requirements are not for one particular localization method or technology, but for the system comprised of many pieces. In addition, the system must meet both 95\% accuracy requirements and safety integrity level requirements in all weather and traffic conditions where operation is intended. Demonstrating the desired integrity levels cannot be proven by vehicle testing alone, where reasonable sized testing fleets would have to be driven for potentially decades to obtain the necessary data~\cite{Kalra2016}. Hence, innovative certification solutions may be necessary~\cite{Kalra2016, Koopman2017}. In addition, the localization requirements presented here are with respect to knowledge of where roads and lanes are in the world. Hence, these requirements are with respect to the map. The map will also have its own uncertainty $\sigma_{map}$ with respect to the global reference, e.g. the WGS-84 datum, as also discussed in~\cite{EDMapConsortium2004}. The relationship between the vehicle's localization uncertainty in the global frame $\sigma_{global}$, its localization uncertainty relative to the map $\sigma_{relative}$, and the uncertainty of the map itself $\sigma_{map}$ with respect to the global frame is given by the following: 
\begin{equation} \label{eq:map-error-global}
\sigma_{global}^2  = \sigma_{relative}^2+ \sigma_{map}^2
\end{equation}
Well geo-referenced maps tied to global datums such as WGS-84 will likely be necessary for interoperability of maps between potentially many map suppliers. These maps can be made with survey-grade equipment and post-processing and could have errors much less than the real-time vehicle localization requirements. 

What has been presented here are localization requirements based on the limiting road geometry. Additional requirements based on operational and other constraints will continue to evolve, but this provides a baseline. These geometrical constraints represent the worst cases; with a-priori highly detailed maps of the environment, the road geometry will be known and hence localization resources can be adjusted on the fly to meet demand and could even be a layer in the map itself. Achieving these requirements represents challenges in sensor and algorithm development along with multi-modal sensor fusion to obtain the reliability levels needed for safe operation. Some techniques involving LiDAR, radar, and cameras rely on a-priori maps and give map-relative position. Others such as GNSS give global absolute position. Combining these and other technologies and selecting those most appropriate for the desired level of autonomous operation in a way that ensures integrity for safe operation is the challenge that lays ahead.

\section*{Acknowledgments}

The authors would like to thank Ford Motor Company for supporting this work.

%\ifCLASSOPTIONcaptionsoff
%  \newpage
%\fi



%\bibliographystyle{IEEEtran}
%\begin{thebibliography}{1}

\bibliographystyle{files/IEEEtran}
\bibliography{files/refs}

%\end{thebibliography}


 %\newpage
\begin{IEEEbiography}[{\includegraphics[width=1in,height=1.25in,clip,keepaspectratio]{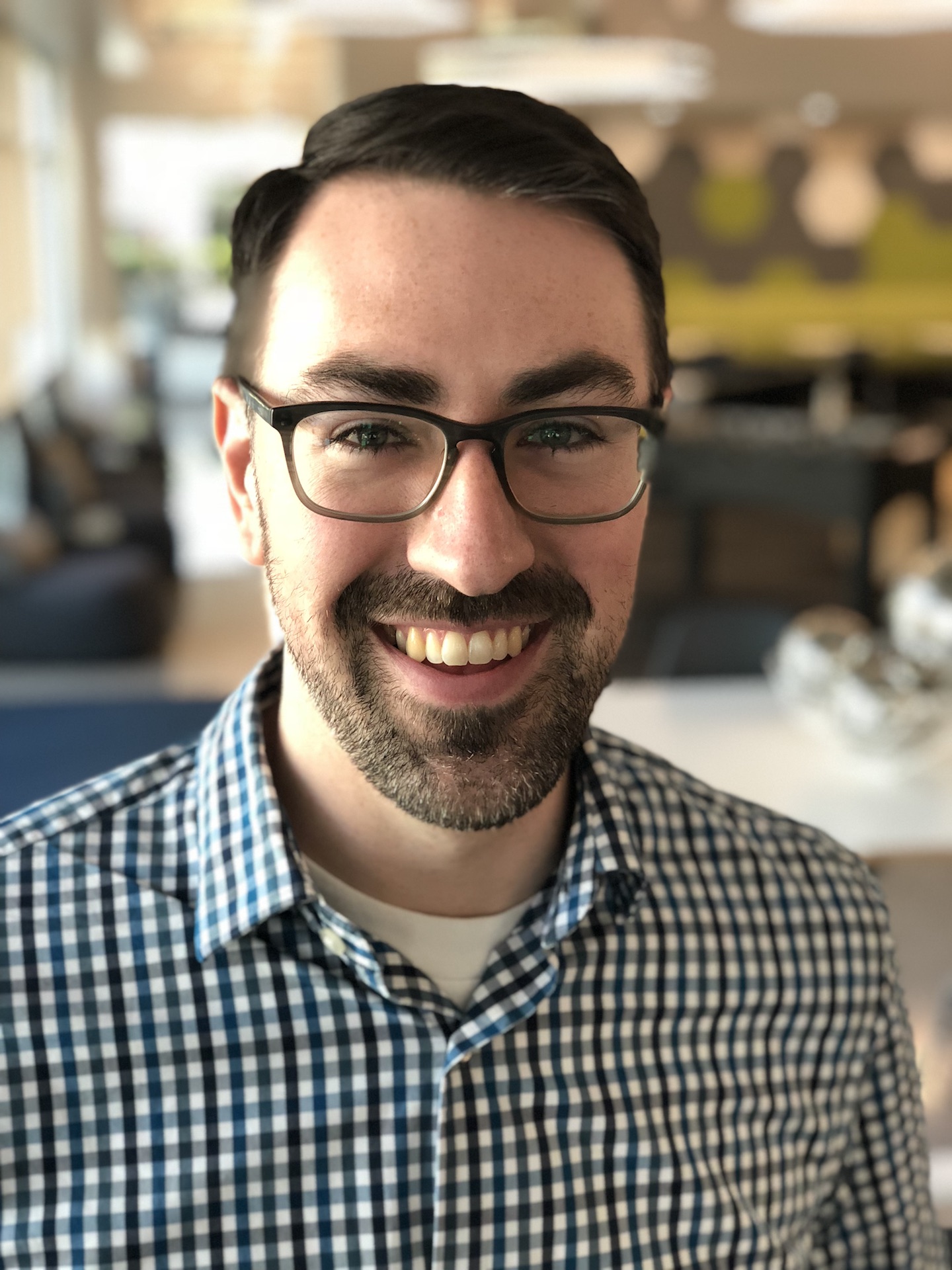}}]{Dr. Tyler G. R. Reid} 
%\begin{IEEEbiography}{Tyler G. R. Reid} 
	is a Research Engineer on the Controls and Automated Systems team at Ford Motor Company working in the area of localization and mapping. He is also a lecturer at Stanford University in Aeronautics and Astronautics. He received his Ph.D. ('17) and M.S. ('12) in Aeronautics and Astronautics from Stanford where he worked in the GPS Research Lab. In 2015, he worked as a Software Engineer at Google's Street View. He completed his B.Eng. in Mechanical Engineering ('10) at McGill University. 
\end{IEEEbiography}
\vskip 0pt plus -1fil
%\begin{IEEEbiography}[{\includegraphics[width=1in,height=1.25in,clip,keepaspectratio]{figs/sarah.jpg}}]{Sarah E. Houts}
\begin{IEEEbiography}[{\includegraphics[width=1in,height=1.25in,clip,keepaspectratio]{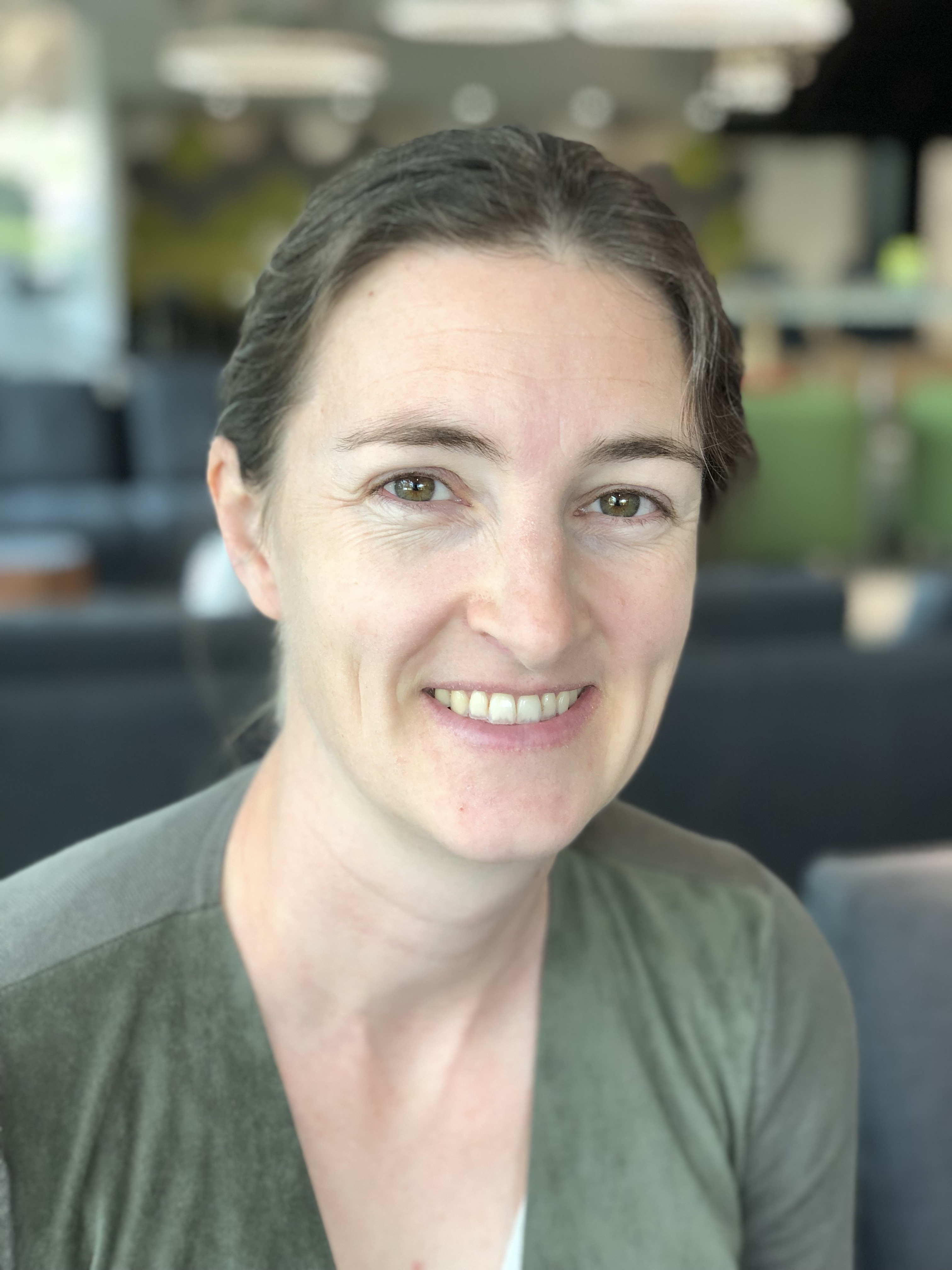}}]{Dr. Sarah E. Houts}
is a Research Engineer at Ford Autonomous Vehicles LLC, working on mapping and localization technologies for Autonomous Vehicles. She received her Ph.D. ('16) and M.Sc. ('08) in Aeronautics and Astronautics from Stanford University, where she worked in the Aerospace Robotics Lab focusing on localization and path planning for autonomous underwater vehicles. She also received her B.S. ('06) in Aerospace Engineering from UC San Diego. 
\end{IEEEbiography}
\vskip 0pt plus -1fil
%\newpage

\begin{IEEEbiographynophoto}{Robert Cammarata}
	is a Supervisor for the Autonomous Vehicle Systems Engineering (AVSE) team at Ford Autonomous Vehicles LLC. One aspect of his current responsibilities include overseeing and developing systems engineering strategies and functional safety activities for autonomous vehicle systems. He achieved a Master of Science in Electrical Engineering from the University of Michigan. Prior to employment at Ford, he held positions at Chrysler, Mercedes Benz RDNA, kVA, Apple, Tesla, and GM developing embedded software, embedded architectures, functional safety analysis and cybersecurity solutions for traditional, hybrid, electrified, and autonomous vehicles.
\end{IEEEbiographynophoto}
\vskip 0pt plus -1fil
\begin{IEEEbiographynophoto}{Dr. Graham Mills} 
	is a Research Engineer at Ford Autonomous Vehicles, LLC specializing in LiDAR calibration and mapping. He received his PhD ('15) in Geomatics and Geology from Queen's University, where his research focused on automated classification of rock surface geometry in LiDAR scans. 
\end{IEEEbiographynophoto}
\vskip 0pt plus -1fil
\begin{IEEEbiography}[{\includegraphics[width=1in,height=1.25in,clip,keepaspectratio]{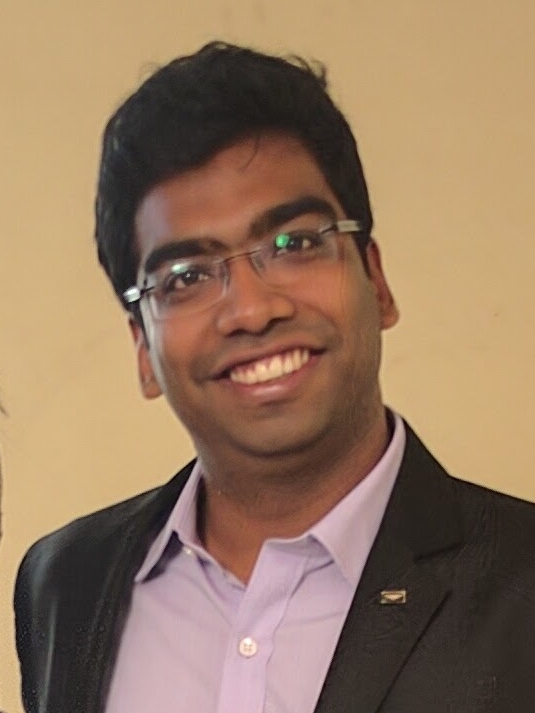}}]{Siddharth Agarwal}
	is a Research Engineer at Ford Autonomous Vehicles LLC. He received his M.S. ('15) in Electrical Engineering from Texas A\&M University with a research focus on Unmanned Aerial Vehicles. His areas of interest include localization, mapping, and multi-agent autonomous systems. 
\end{IEEEbiography}
\vskip 0pt plus -1fil
\begin{IEEEbiography}[{\includegraphics[width=1in,height=1.25in,clip,keepaspectratio]{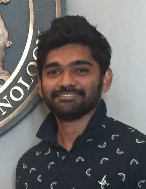}}]{Ankit Vora}
	Ankit Vora is a Research Engineer at Ford Autonomous Vehicles LLC working in the area of mapping and localization. His research focus is on SLAM, state estimation for autonomous vehicles, localization, and place recognition. He received his M.S.E. ('16) in Mechanical Engineering, specializing in Robotics, from the University of Pennsylvania.
\end{IEEEbiography}
\vskip 0pt plus -1fil
\begin{IEEEbiography}[{\includegraphics[width=1in,height=1.25in,clip,keepaspectratio]{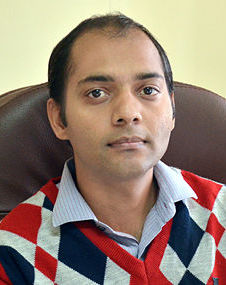}}]{Dr. Gaurav Pandey} is a Technical Expert in the Controls and Automated Systems department of Ford Motor Company. He is currently leading the mapping and localization group at Ford and is working on developing localization algorithms for SAE level 3 and level 4 autonomous vehicles. Prior to Ford, Dr. Pandey was an Assistant Professor at the Electrical engineering department of Indian Institute of Technology (IIT) Kanpur in India. At IIT Kanpur he was part of two research groups (i) Control and Automation, (ii) Signal Processing and Communication. His research focus is on visual perception for autonomous vehicles and mobile robots using tools from computer vision, machine learning, and information theory. He did is B-Tech from IIT Roorkee in 2006 and completed his Ph.D. from University of Michigan, Ann Arbor in December 2013.
\end{IEEEbiography}

%
%\begin{IEEEbiographynophoto}{Gaurav Pandey}
%	Biography text here.


%\vfill


\end{document}